\documentclass[accepted]{melba}
\usepackage{mwe} % to get dummy images, only for the example
\usepackage{amsmath,amsfonts}
\usepackage{multirow}
\usepackage{enumitem}
\usepackage[table]{xcolor}
\usepackage[dvipsnames]{xcolor}
\usepackage{adjustbox}
\usepackage{pifont}
\usepackage{xurl}
\usepackage{hyperref}
\usepackage{tikz}
\usetikzlibrary{tikzmark, calc}
\usepackage{forest}
\usepackage[T1]{fontenc}
\usepackage{amsmath}
\usepackage{amssymb}
\usepackage{mathtools}
\usepackage[capitalize,noabbrev]{cleveref}
\usepackage{tabularx}
\usepackage{breakurl}
\usepackage{pifont}
\usepackage{placeins}

\melbaid{2026:018}  % This is provided upon by the publishing editor
\doi{10.59275/j.melba.2026-87e3}
\melbaauthors{Di Piazza, Lazarus, Nempont and Boussel}  % Note: this one is also used to set the pdf 'authors' metadata
\email{theo.dipiazza@creatis.insa-lyon.fr}
\volume{2026}
\firstpageno{359}  % Communicated by the publishing editor
\melbayear{2026}  % The publication year
\datesubmitted{2025-10-27}  % Date submitted to MELBA: mm/yyyy
\datepublished{2026-06-15}  % Today's date: mm/yyyy

\ShortHeadings{CT-SSG}{Di Piazza, Lazarus, Nempont and Boussel}

\title{Structured Spectral Graph Representation Learning for Multi-label Abnormality Analysis from 3D CT Scans}
\author{
	\firstname Theo \surname Di Piazza\aff{1,2},
	\firstname Carole \surname Lazarus\aff{3},
    \firstname Olivier \surname Nempont\aff{3},
    \firstname Loic \surname Boussel\aff{1, 2}
}
\affiliations{% <- trailing '%' to avoid unwanted indent
	\num 1 \addr INSA Lyon, University of Lyon, CNRS, INSERM, CREATIS UMR 5220, U1294, Villeurbanne, France \\
	\num 2 \addr Hospices Civil de Lyon, Lyon, France \\
	\num 3 \addr Philips Clinical Informatics, Innovation Paris, France
}

\abstract{%   <- trailing '%' for backward compatibility of .sty file
    With the growing volume of CT examinations, there is an increasing demand for automated tools such as organ segmentation, abnormality detection, and report generation to support radiologists in managing their clinical workload. Multi-label classification of 3D Chest CT scans remains a critical yet challenging problem due to the complex spatial relationships inherent in volumetric data and the wide variability of abnormalities. Existing methods based on 3D convolutional neural networks struggle to capture long-range dependencies, while Vision Transformers often require extensive pre-training on large-scale, domain-specific datasets to perform competitively. In this work, we propose a 2.5D alternative by introducing a new graph-based framework that represents 3D CT volumes as structured graphs, where axial slice triplets serve as nodes processed through spectral graph convolution, enabling the model to reason over inter-slice dependencies while maintaining complexity compatible with clinical deployment. Our method, trained and evaluated on 3 datasets from independent institutions, achieves strong cross-dataset generalization, and shows competitive performance compared to state-of-the-art visual encoders. We further conduct comprehensive ablation studies to evaluate the impact of various aggregation strategies, edge-weighting schemes, and graph connectivity patterns. Additionally, we demonstrate the broader applicability of our approach through transfer experiments on automated radiology report generation and abdominal CT data.
    This work extends our previous contribution presented at the MICCAI 2025 EMERGE Workshop.
    A project page is available at \url{https://theodpzz.github.io/projects/ctssg/}.
    }


\begin{document}

% top matter
\twocolumn[\maketitle]
% comment the preceedings and uncomment the following if the authors list + abstract is longer than one page
% \maketitle
% \twocolumn

%%%%%%%%%%%%%%%%%%%%%%%%%%%%%%%%%%%%%%%%%%%%%%%%%%%%%%%%%%%%%%%%%%%%%%%%%%%
% INTRODUCTION
%%%%%%%%%%%%%%%%%%%%%%%%%%%%%%%%%%%%%%%%%%%%%%%%%%%%%%%%%%%%%%%%%%%%%%%%%%%
\section{Introduction}
Computed Tomography (CT) is a cornerstone imaging modality in clinical practice, providing radiologists with detailed three-dimensional views of the thorax and enabling the accurate detection of a wide range of abnormalities~\citep{patel_ct_2024}. However, the increasing volume of chest CT scans poses significant challenges for radiologists, who face mounting demands and time constraints~\citep{broder_increasing_2006}. This has created an urgent need for automated systems capable of assisting healthcare professionals to manage their increasing workload~\citep{najjar_redefining_2023}.

In medical imaging, early developments in automated abnormality detection predominantly focused on 2D modalities, facilitated by the availability of large-scale datasets such as CheXpert~\citep{irvin_chexpert_2019} and MIMIC-CXR~\citep{johnson_mimic-cxr_2019}. Early work on 3D chest CT abnormality classification initially addressed single-label classification, targeting one abnormality at a time~\citep{panwar_deep_2020}. Yet, multi-label abnormality classification is of paramount importance for clinical decision support, as it allows simultaneous detection of multiple co-occurring abnormalities and leverages inter-abnormality relationships to improve diagnostic performance~\citep{draelos_machine-learning-based_2021}. Moreover, multi-label classification serves as a versatile pretext task that can later be fine-tuned for more specialized objectives, such as report generation or disease progression modeling~\citep{tanida_interactive_2023}.

\begin{figure}[h]
    \centering
    \includegraphics[width=\columnwidth]{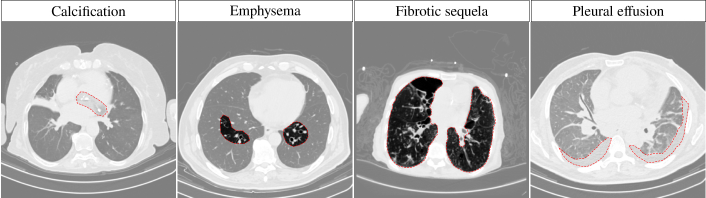}
    \caption{Axial slices from 3D CT Scans from \texttt{CT-RATE}, with abnormalities manually contoured in red, illustrating distinct visual characteristics.}
    \label{fig:example_anomalies}
\end{figure}

Despite its clinical relevance, multi-label abnormality classification in 3D chest CT remains a highly challenging task due to the broad diversity of abnormalities, as illustrated in Figure~\ref{fig:example_anomalies}. Furthermore, the volumetric nature of CT data necessitates the development of computationally efficient architectures that are scalable and suitable for real-world clinical deployment~\citep{aravazhi_integration_2025}.

Early approaches to multi-label abnormality classification in CT imaging predominantly leveraged fully convolutional networks. The recent release of {\tt CT-RATE}, a large-scale public dataset~\citep{hamamci_generalist_2026} containing chest CT scans from over 21,000 unique patients paired with radiology reports, has significantly broadened the scope of CT-based research. This includes tasks such as synthetic volume generation~\citep{hamamci_generatect_2023} and automatic report generation~\citep{hamamci_ct2rep_2024}. Notably, many of these methods adopt visual encoder architectures based on video vision transformers~\citep{arnab_vivit_2021}, which model the 3D CT Scan as a set of 3D patches. However, 3D Transformers-based methods often rely on extensive pre-training on large, domain-specific datasets to achieve competitive performance~\citep{hamamci_generatect_2023}. Recently, prior work has empiricaly demonstrated that 2.5D modeling, representing a CT volume as a set of slices rather than a set of 3D patches, can outperform purely 3D approaches. For instance, CT-Net introduced a 2.5D alternative to full 3D CNNs by modeling CT volumes as sequences of axial slices, processed independently using a 2D backbone~\citep{draelos_machine-learning-based_2021}. While CT-Net demonstrated strong performance on 83 abnormalities within an internal dataset, its generalization across datasets and tasks remained unexplored, largely due to the scarcity of publicly available annotated CT datasets at the time.

Prior works describe graphs as "the main modality of data we receive from nature"~\citep{velickovic_everything_2023}. From this perspective, most machine learning applications can be seen as special cases of graph representation learning, including Transformers~\citep{joshi_transformers_2025}, which has lead to significant efforts in recent years across various domains of application~\citep{zhou_graph_2020}. Specifically, Transformers operate on fully connected graphs, where attention mechanisms learn adaptive edge weights between all node pairs~\citep{giovanni_over-squashing_2023}. While this formulation has proven expressive, it entails dense connectivity~\citep{fey_fast_2019}, require extensive pre-training~\citep{bommasani_opportunities_2022} and is useful for tasks where we do not have an apriori graph structure~\citep{jumper_highly_2021}, which may be suboptimal for modeling localized spatial dependencies inherent to 3D medical volumes. In contrast, representing 3D CT scans as structured graphs provides a more flexible and physically grounded framework: it allows explicit control over neighborhood definitions, edge weighting strategies, and hierarchical topologies. Recent advances in Graph Neural Networks (GNN) have demonstrated their ability to model complex relational structures across diverse imaging modalities~\citep{ahmedt-aristizabal_graph-based_2021}. In medical imaging, GNNs have been successfully applied to tasks such as automated report generation~\citep{liu_exploring_2021}, where they capture semantic dependencies among knowledge entities, and whole-slide image analysis, where hierarchical graph representations enhance abnormality classification~\citep{guo_higt_2023}. These successes suggest that GNNs hold strong potential for extending 3D modeling paradigms in chest CT analysis, particularly in scenarios where volumetric context and inter-slice dependencies are critical.

Building on the representational flexibility of 2.5D approaches and the relational expressiveness of graph neural networks, we introduce CT-SSG (\textbf{S}tructural \textbf{S}pectral \textbf{G}raph for \textbf{C}omputed \textbf{T}omography), a framework that formally represents 3D CT volumes as structured graphs. In this formulation, each node corresponds to a triplet of axial slices, and edges encode spatial dependencies parameterized by inter-slice spacing along the z-axis. Slice-level features interact through spectral-domain graph convolutions, enabling efficient modeling of both local anatomical context and global volumetric structure. Spatial awareness is further reinforced through an axial positional embedding. We conduct extensive experiments to analyze the impact of graph topology, edge weighting, and feature aggregation strategies, comparing CT-SSG with both standard neural encoders and domain-specific CT architectures. Comprehensive ablations and transfer studies demonstrate the generality of our formulation, including applications to automated radiology report generation and cross-organ adaptation to abdominal CT scans for multi-label abnormality classification.

To summarize our contributions and key advantages of our work: (1) \textbf{CT-SSG}: We propose CT-SSG, a new visual encoder that models a 3D CT volume as a graph of triplet axial slices. To capture spatial dependencies, we introduce a Triplet Axial Slice Positional Embedding, along with an edge-weighting strategy for relative position awareness within a spectral-domain GNN module; (2) \textbf{Cross-dataset generalization}: CT-SSG demonstrates strong cross-dataset generalization, maintaining consistent performance when trained on a public Turkish dataset and evaluated on independent datasets from the United States and France. Additionally, we demonstrate the transferability of CT-SSG's pretrained weights from chest to abdominal CT scans, highlighting its potential as a versatile backbone for a broad range of 3D medical imaging tasks; (3) \textbf{Ablation study}: We conduct thorough ablation studies to analyze the impact of model depth, hyperparameter choices, graph topology, and connectivity patterns across different convolutional operators. Additionally, we evaluate models under patient-specific variations, including z-axis translations and voxel intensity perturbations. (4) \textbf{Transfer to Report generation}: Beyond multi-label abnormality classification, we evaluate CT-SSG on automated radiology report generation, demonstrating that the learned representations are transferable and effective for related CT-based downstream tasks. (5) \textbf{Transfer to Abdominal CT for Abnormality Classification}: Although our primary focus is chest CT, we evaluate CT-SSG representations via linear probing on abdominal CT in a low-data regime. We find that chest-pretrained backbones yield stronger performance than supervised training from scratch when fewer than 3,750 samples are available, highlighting the transferability of our approach across anatomical domains.

\begin{figure*}[!ht]
	\centering
	\includegraphics[width=1.00\textwidth]{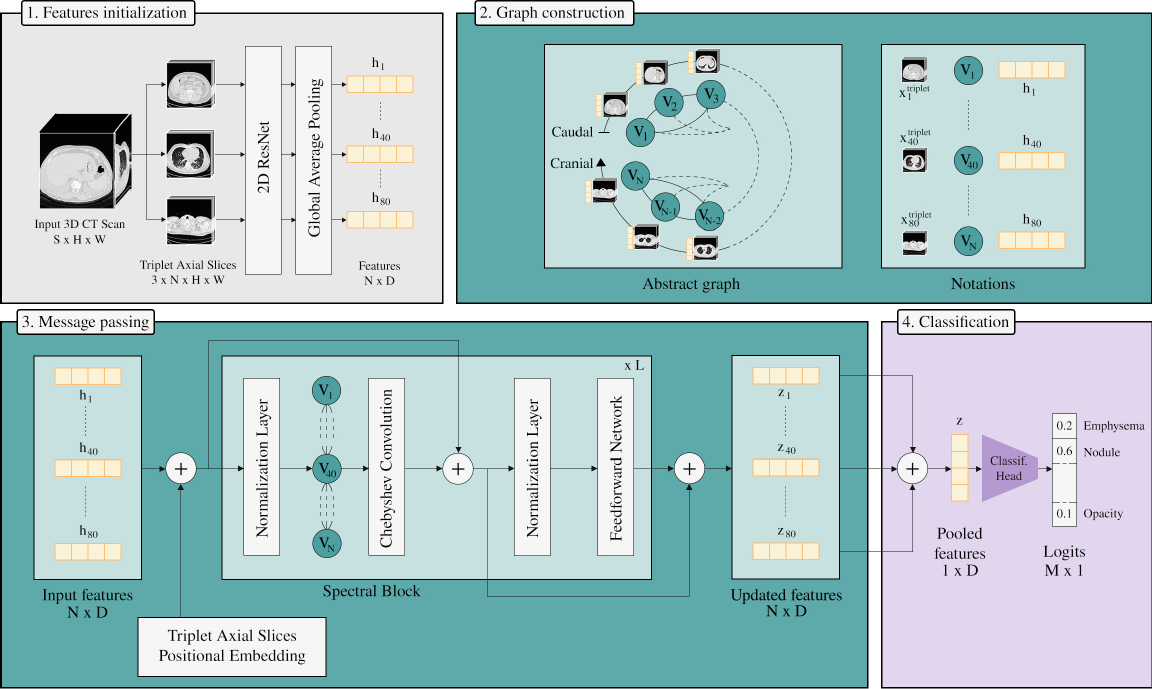}
	\caption{CT-SSG Architecture Overview. Adjacent axial slices are grouped into triplets, each representing a node in a graph. Edges between nodes are weighted according to their physical distance along the z-axis. Node features are enhanced with Triplet Axial Slices positional embeddings, and then processed by a Spectral Block that incorporates Chebyshev graph convolution for structured spectral modeling. The resulting node representations are aggregated via mean pooling and passed to a classification head to predict abnormalities.}
    \label{fig:overview}
\end{figure*}

%%%%%%%%%%%%%%%%%%%%%%%%%%%%%%%%%%%%%%%%%%%%%%%%%%%%%%%%%%%%%%%%%%%%%%%%%%%
% Related works
%%%%%%%%%%%%%%%%%%%%%%%%%%%%%%%%%%%%%%%%%%%%%%%%%%%%%%%%%%%%%%%%%%%%%%%%%%%
\section{Related Works}

\subsection{3D Visual Encoder}
\label{rw_visual_encoder}
\textbf{3D Convolutional Neural Network.} Early advances in both 2D and 3D imaging, spanning natural images and medical modalities, have been predominantly driven by Convolutional Neural Networks (CNNs), which demonstrated strong capabilities in extracting fine-grained visual features~\citep{lecun_deep_2015}. CNN-based architectures have been successfully applied to a wide range of tasks, including segmentation~\citep{ilesanmi_reviewing_2024}, classification~\citep{he_deep_2015}, and image captioning~\citep{kougia_survey_2019}, across diverse domains such as medical imaging~\citep{anaya-isaza_overview_2021}, earth observation~\citep{bianchi_uav_2021}, and sports analytics~\citep{chang_basketball_2024}.

\textbf{3D Transformer Neural Network.} Despite their success, CNNs inherently struggle to capture long-range dependencies due to their limited receptive fields, which can hinder their ability to model contextual information, an essential requirement in 3D imaging for understanding large-scale anatomical structures~\citep{ma_u-mamba_2024}. Inspired by breakthroughs in Natural Language Processing~\citep{devlin_bert_2018}, Vision Transformers (ViTs) were introduced for 2D visual modalities, offering an alternative that leverages self-attention mechanisms~\citep{vaswani_attention_2023} to model global context by enabling interactions between distant image regions~\citep{dosovitskiy_image_2021}.
Building upon these principles, Vision Transformers have been extended to 3D data, including applications in video analysis and 3D medical imaging~\citep{wang_review_2023}. Notably, ViViT, an adaptation of the Vision Transformer for video sequences, applies a Spatial Transformer to model interactions among spatial tokens for each temporal step, followed by a Temporal Transformer to capture dependencies along the temporal axis~\citep{arnab_vivit_2021}. In the context of CT imaging, ViViT has further inspired frameworks for synthetic volume generation~\citep{hamamci_generatect_2023} and automated clinical report synthesis~\citep{hamamci_ct2rep_2024}. Similarly, the Swin Transformer, initially designed for 2D vision tasks~\citep{liu_swin_2021}, introduces a hierarchical architecture with shifted windows that enables local and global context modeling while efficiently handling large variations in the scale of visual entities. Swin Transformer was adapted to 3D modalities for various tasks such as video understanding~\citep{liu_video_2021}, indoor scene understanding~\citep{yang_swin3d_2023} and organs segmentation of 3D medical images~\citep{tang_self-supervised_2022}.

\textbf{2.5D Neural Network.} While Vision Transformers excel at modeling long-range dependencies, they often require extensive pre-training on large-scale, domain-specific datasets to achieve competitive performance, a limitation in medical imaging where annotated datasets are comparatively scarce~\citep{hamamci_generatect_2023}. A widely adopted alternative is transfer learning from models pre-trained on large-scale natural image datasets~\citep{zhang_adapting_2023}. In 3D Chest CT imaging, CT-Net was among the first approaches to propose a 2.5D strategy, representing volumetric CT data as stacks of axial slices~\citep{draelos_machine-learning-based_2021}. Feature maps are extracted from each slice using a 2D image encoder pre-trained on natural images, then aggregated through a lightweight 3D convolutional network to produce a compact volumetric representation. This idea was further extended by CT-Scroll, which introduced a hybrid scheme wherein the volume is represented as a set of visual tokens, each associated with triplets of slices~\citep{di_piazza_imitating_2025}. These tokens interact through attention mechanisms and are subsequently aggregated via mean pooling. While 3D approaches, such as ViTs or Swin Transformers, incorporate spatial awareness through positional embeddings~\citep{dosovitskiy_image_2021} or relative positional biases~\citep{liu_swin_2021}, 2.5D methods lack explicit or implicit modeling of spatial continuity within the volume. This limitation may constrain their ability to effectively capture both short- and long-range spatial dependencies.

\subsection{Foundation Models for CT Imaging} \label{sec:related:foundation}

Recently, large-scale foundation models have emerged as powerful feature extractors for medical imaging, including CT volumes. Approaches such as CT-CLIP~\citep{hamamci_generalist_2026}, CT-FM~\citep{pai_vision_2025}, Merlin~\citep{blankemeier_merlin_2024} and COLIPRI~\citep{wald_comprehensive_2026} leverage large-scale pretraining to learn transferable representations that perform well when transfered to more specific task under limited fine-tuning. Such frameworks often rely on self-supervised objectives including visual reconstruction~\citep{hamamci_generatect_2023}, visual contrastive learning~\citep{pai_vision_2025} or vision-language alignment~\citep{wald_comprehensive_2026,blankemeier_merlin_2024,hamamci_generalist_2026} from large-scale datasets. Similarly, general-purpose self-supervised models such as DINOv2~\citep{oquab_dinov2_2024}, that trains a 1B parameters Vision Transformer model on large-scale curated data from diverse sources, have been adapted to medical images to obtain strong slice-level embeddings, both for CT and MRI~\citep{muller-franzes_medical_2025}, as well as for multi-modal patch embeddings from multi-sequence MRI~\citep{scholz_mm-dinov2_2025}. While these methods excel at learning generic visual representations, the visual encoders typically operate within Euclidean framework and do not explicitly model structured inter-slice relationships. In contrast, our work focuses on structured representation learning for CT volumes, a paradigm which can be complementary to foundation models. Notably, CT-SSG can incorporate features extracted from any 2D pretrained encoder, while providing a structured graph reasoning layer tailored to volumetric medical data.

\subsection{Graph Neural Network}
\label{rw_gnn}
In various application domains such as biology~\citep{reiser_graph_2022} or transportation~\citep{makarov_graph_2024}, graphs are a common representation of data found in nature~\citep{velickovic_everything_2023}. A graph, denoted as $\mathcal{G} = \{\mathcal{V}, \mathcal{E}\}$ consists of a set of edges $\mathcal{E}$ which model the connections between a set of nodes $\mathcal{V}$. In deep learning, GNNs have become the main approach for tasks involving graph-structured data~\citep{bechler-speicher_intelligible_2024}, where each node is associated with a vector representation, which is iteratively updated through neighborhood aggregation during the forward message passing process. 

Representative models mainly include Convolutional GNNs (GraphConv), which aggregate neighboring node features through graph-based convolutions~\citep{defferrard_convolutional_2017} or Attentional GNNs (GAT), which leverage attention mechanisms to weight the importance of different neighbors during aggregation~\citep{velickovic_graph_2018}. Inspired by the attention mechanism~\citep{bahdanau_neural_2016} and self-attention mechanism of the Transformer~\citep{vaswani_attention_2023}, the motivation of Graph Attention is to compute a representation of every node as a weighted average of its neighbors~\citep{brody_how_2022}. While spatial networks, including Graph Convolution and Graph Attention, define graph convolution as a localized averaging operation with learned weights, spectral networks define convolution via eigen-decomposition of the graph Laplacian~\citep{zhang_magnet_2021}. In such spectral networks, the convolution operator is defined in the Fourier domain through localized spectral filters.

In medical imaging, GNNs have been used in tasks such as medical knowledge integration in 2D X-ray radiology report generation~\citep{liu_exploring_2021} to incorporate prior knowledge as a graph of connected textual medical concepts, and Whole Slide Image (WSI) analysis~\citep{guo_higt_2023} to model the hierarchical structure of the pyramids WSI. In the context of Computed Tomography, recent work~\citep{kalisch_ct-graph_2025} models CT volumes as graphs by grouping patches based on anatomical segmentation for report generation. In contrast, our method is purely data-driven, requiring no segmentation labels and is therefore applicable in settings without anatomical annotations. For clarity, this study is restricted to segmentation-free paradigms, differentiating it from anatomical segmentation-based graph methods. Separately, multi-view graph representations have been explored in 3D medical imaging, where each node encodes features from orthogonal axial, sagittal, and coronal slices using a frozen 2D Vision Transformer~\citep{kiechle_graph_2024}.

Building on the empirical success of 2.5D approaches, we propose a principled formulation of 3D CT volumes as structured graphs of axial slices. This perspective unifies slice-level representations with inter-slice dependencies, enabling systematic investigation of graph topologies, edge-weighting schemes, and aggregation mechanisms. This work formalizes CT interpretation within a graph-based framework, providing a flexible and general paradigm that bridges 2D and volumetric modeling.

%%%%%%%%%%%%%%%%%%%%%%%%%%%%%%%%%%%%%%%%%%%%%%%%%%%%%%%%%%%%%%%%%%%%%%%%%%%
% METHOD
%%%%%%%%%%%%%%%%%%%%%%%%%%%%%%%%%%%%%%%%%%%%%%%%%%%%%%%%%%%%%%%%%%%%%%%%%%%
\section{Method}
\label{sec:method}

As shown in Figure~\ref{fig:overview}, CT-SSG models the 3D CT Scan as a graph of \textit{triplet axial CT slices} with undirected edges weighted by their \textit{physical distance along the caudal-cranial axis}. Node features interact through a spectral domain module, before being pooled and given to a classification head to predict abnormalities. A comprehensive PyTorch pseudocode table outlining each operation, its semantic role, and corresponding tensor shapes is provided in Appendix~\ref{appendix:table:pseudo_code}.

\begin{table}[h]
\centering
\begin{adjustbox}{width=1.00\columnwidth}
\begin{tabular}{c c c}
\toprule
Symbol & Description & Value\\
\toprule
% S
$S$ & 
Number of axial slices & 
$240$\\

% H
\cellcolor[gray]{0.925}$H_{s}$ &
\cellcolor[gray]{0.925}Height of an axial slice &
\cellcolor[gray]{0.925}$480$\\

% W_s
$W_{s}$ & 
Width of an axial slice & 
$480$\\

% N
\cellcolor[gray]{0.925}$N$ & 
\cellcolor[gray]{0.925}Number of triplet axial slices &
\cellcolor[gray]{0.925}$80$\\

% C
$C$ & 
Number of axial slices per triplet & 
$3$\\

% d
\cellcolor[gray]{0.925}$d$ & 
\cellcolor[gray]{0.925}Dimension of latent space &
\cellcolor[gray]{0.925}$512$\\

% L
$L^{*}$ & 
Spectral module depth &
$1$\\

\cellcolor[gray]{0.925}$q_l^{*}$ & 
\cellcolor[gray]{0.925}Receptive field for layer $l$ &
\cellcolor[gray]{0.925}$16$\\

% K
$K^{*}$ & 
Chebyshev filter size &
$3$\\
\toprule
\end{tabular}
\end{adjustbox}
\caption{Summary of key notations and optimal experimental settings. Symbols marked with * denote tuned hyperparameters.}
\label{table:notations}
\end{table}

\subsection{Notations}

We consider a multi-label abnormality classification task with an input space $\mathcal{X} \in \mathbb{R}^{S \times H_{s} \times W_{s}}$ and a target space $\mathcal{Y} \in [1, \cdots, M]$. $S$ refers to the number of axial slices, each of dimension $H_{s} \times W_{s}$. $M$ is the number of abnormalities. Table~\ref{table:notations} details each variable with description and corresponding value for experiments.

\subsection{Features Initialization}
\label{features_init}

Following a 2.5D strategy, we partition the input volume $X$ into $N$ non-overlapping triplets of slices to encode minimal local 3D context while maintaining compatibility with 2D backbones, resulting in a tensor of shape $N\times C \times H_{s} \times W_{s}$. Using three adjacent slices allows the network to capture inter-slice anatomical continuity and subtle volumetric context without resorting to full 3D convolutions or excessively large context windows that may dilute localized patterns. Each triplet, noted as $x_{i}^{\text{triplet}} (i \in \{1, \ldots, N\})$, is processed independently by a learnable 2D ResNet-18 backbone~\citep{he_deep_2015}, extracting spatial features. The $N$ output features maps are subsequently aggregated via mean pooling to produce a compact representation for each slice triplet. This features initialization step maps each triplet of axial slice into a $d$-dimensional embedding, denoted as $\bar{h}_{i} \in \mathbb{R}^{d}$. The use of ResNet-18 provides a favorable trade-off between representational capacity and computational efficiency while allowing us to focus on evaluating the contribution of the proposed graph-based feature aggregation. Importantly, the proposed CT-SSG formulation is backbone-agnostic. To empirically validate this property, we additionally evaluate a higher-capacity ResNet-34 encoder in Appendix~\ref{sec:appendix:scaling}, demonstrating consistent performance scaling without modification to the graph architecture.

\subsection{Triplet Axial Slices Positional Embeddings}
\label{sec:tfcol}

After obtaining all triplet slices embeddings $\bar{H} = [\bar{h}_{1}, \ldots, \bar{h}_{N}]$, triplet axial slices positional embeddings are added to retain positional information along the caudal-cranial axis. We use learnable 1D position embeddings, denoted as $P^{\text{axial}}_{\text{pos}} \in \mathbb{R}^{N \times d}$, resulting as a sequence of embedding vectors ${H}$, such that:
\begin{equation}
    {H} = \bar{H} + P^{\text{axial}}_{\text{pos}} \, .
\end{equation}

\subsection{Graph Construction} \label{sec:tfrow}

We define the volumetric representation as a graph $\mathcal{G} = (\mathcal{V}, \mathcal{E}, {H}, A)$. In this section, we define nodes, edges, node features and the adjacency matrix.

\paragraph{Nodes.} $\mathcal{V} = \{ v_i \}_{i=1}^{N}$ is the set of nodes, where each node $v_i$ represents a triplet of 3 consecutive axial slices.

\paragraph{Edges.} $\mathcal{E} \subseteq \mathcal{V} \times \mathcal{V}$ is the set of edges, where an edge $(v_i, v_j) \in \mathcal{E}$ is weighted based on a function of inter-triplet distance and z-axis spacing. The weight of an edge $(v_i, v_j)$, denoted as $w_{i, j} \in \mathbb{R}^{+}$, is defined such that:
\begin{equation} \label{edge_weight_fun}
    w_{i, j} = 1 + \frac{1}{1+3 \times |i-j| \times s_{z}} \, ,
\end{equation}

where $s_{z}$ is the spacing along the caudal-cranial axis in millimeters.

We further investigate the impact of graph connectivity by exploring a family of topologies parameterized by a receptive field size $q \in \mathbb{N}^{+}$. Specifically, we construct an undirected edge $(v_i, v_j) \in \mathcal{E}$ between nodes if their corresponding triplet slices are at most $q$ steps apart in the sequence, yielding the edge set:
\begin{equation}
    \mathcal{E} =  \{ (v_i, v_j) \ | \ |i-j| \leq q\}\, .
\end{equation}

In Section~\ref{res_ablation}, we perform a comprehensive ablation study to assess how varying $q$ influences the performance of different GNN architectures, highlighting the role of graph receptive field in modeling caudal-cranial axis dependencies within 3D CT volumes.

\paragraph{Nodes features.} ${H} = \{{h}_{1}, \ldots, {h}_{N} \} \in \mathbb{R}^{N \times d}$ is the node feature matrix, where ${h}_i \in \mathbb{R}^{d}$ denotes the feature embedding of node $v_i$.

\paragraph{Adjacency matrix.} $A \in \mathbb{R}^{N \times N}$ is the weighted adjacency matrix, where $A_{ij} = w_{i, j} \in \mathbb{R}^{+}$ encodes the connectivity and spatial relationship between triplets, $w_{i, j}$ being the edge weight such that:  
\begin{equation}
    A_{ij} =
    \begin{cases}
        w_{ij}, & \text{if } (v_i, v_j) \in \mathcal{E} \\
        0, & \text{otherwise.}
    \end{cases} \,
\end{equation}

\begin{figure}[h]
    \centering
    \includegraphics[width=1.00\columnwidth]{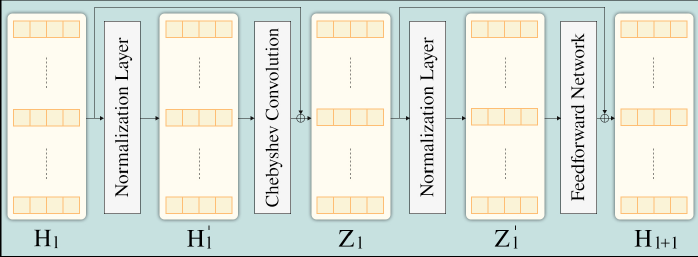}
    \caption{Spectral Block with detailed notations. Input features are given to a first normalization layer, followed by spectral graph convolutions with a residual skip connection. These updated features are then fed to a feedforward neural network followed by a second normalization layer with a residual skip connection.}
    \label{fig:spectral_block}
\end{figure}

\subsection{Spectral Domain Module}

A key challenge in this formulation is the variability in anatomical positioning across patients due to differences in scan length and body proportions. Traditional spatial graph convolutions, such as GraphConv~\citep{morris_weisfeiler_2021}, aggregate information from fixed local neighborhoods, which can be suboptimal in this context as anatomical structures do not consistently align across scans. Instead, we leverage Chebyshev convolutions~\citep{defferrard_convolutional_2017} to define graph convolutions in the spectral domain, each followed by a feedforward neural network. Unlike spatial approaches, which struggle with non-uniform neighborhood structures~\citep{bruna_spectral_2014}, ChebConv utilizes polynomial approximations of the graph Laplacian~\citep{belkin_laplacian_2001} to capture hierarchical feature representations while preserving spatial localization. This allows the model to adapt to variations in caudal-cranial slice positioning and effectively learn long-range anatomical relationships, making it more robust to inter-patient variability. 

We introduce a Spectral Module, denoted as $\Phi_{\text{SM}}$, consisting $L$ Spectral Blocks. Each block consists of two sublayers. While Figure \ref{fig:overview} shows the overall CT-SSG architecture, Figure~\ref{fig:spectral_block} presents a detailed schematic of the spectral block, where all operations and symbols are explicitly annotated to facilitate interpretation of the notation.

The first sublayer consists of a Normalization Layer~\citep{ba_layer_2016}, noted $f_{l}^{\text{LN}}$, and followed by a spectral convolution. Specifically, we leverage a Chebyshev Convolution, denoted as $f_{l}^{\text{Cheb}}$, to benefit from its polynomial formulation that allows us to capture information from $K$-hop neighborhood. Let ${H}_{0}=H$, and ${H}_{l}$ denote the input features of the $l$-th block, we formaly define the forward pass in the first sublayer such that:
\begin{equation}
    Z_{l} = {H}_{l} + \left(f_{l}^{\text{Cheb}} \circ f_{l}^{\text{LN}} \right)({H}_{l}) \, .
\end{equation}

For the Chebyshev convolution, denoted as $f_{l}^{\text{Cheb}}$, the scaled and normalized Laplacian $\hat{L}$ is defined as:
\begin{equation}  
\hat{L} =\frac{2}{\lambda_{\text{max}}} (D - A) - I \, ,
\end{equation}  

where $\lambda_{\text{max}}$ is the largest eigenvalue of the graph Laplacian $L = D - A$. The degree matrix $D$ is a diagonal matrix where $D_{i, i} = \sum_{j=1}^{N} w_{i, j}$. The convolution operation is parameterized using Chebyshev polynomials $T_{j}(\hat{L}) \in \mathbb{R}^{N \times N}$, resulting in a recurrence relation for the transformation of the node feature matrix. Let $\theta_{l,k} \in \mathbb{R}^{d \times d}$ be the learnable parameters, and \( K \) be the Chebyshev filter size. The recurrence relation is given by:
\begin{equation}  
f_{l}^{\text{Cheb}}(X) = \sum_{k=0}^{K-1} T_{l,k}(\hat{L}) X \theta_{l,k} \, .
\label{equation:cheb}
\end{equation}  

We investigate the effect of the filter size $K$ on model performance in the ablation study presented in Section~\ref{res_ablation}. The Chebyshev convolution was implemented with \href{https://pytorch-geometric.readthedocs.io/en/2.5.0/generated/torch_geometric.nn.conv.ChebConv.html}{ChebConv module} from \href{https://pytorch-geometric.readthedocs.io/en/2.5.0/index.html}{PyTorch Geometric}. Spectral graph convolutions are defined with respect to the graph Laplacian and are therefore tied to the underlying graph topology. In our setting, this dependency is mitigated by the use of a fixed and regular graph structure reflecting the axial organization of CT volumes, which is shared across all samples. We note that extending the proposed approach to substantially different or data-dependent graph topologies may require adapted spectral formulations.

The second sublayer consists of another Normalization Layer, denoted as $g_{l}^{\text{LN}}$, followed by a feedforward neural network, noted $g_{l}^{\text{FNN}}$ and implemented as a linear layer followed by a GELU activation function~\citep{shazeer_glu_2020}. The second sublayer is also followed by a residual connection, as followed:
\begin{equation}  
{H}^{l+1} = Z^{l} + \left(g_{l}^{\text{FFN}} \circ g_{l}^{\text{LN}} \right)(Z^{l}) \, .
\end{equation} 

Formally, the Spectral Module outputs updated features, denoted as ${Z} = {H}_{L} = [{z}_{1}, \ldots, {z}_{N}]$ with $z_{i} \in \mathbb{R}^{d}$ being the updated features for the $i$-th node, such that:
\begin{equation}
    {Z} = \Phi_{\text{SM}}({H}) \, .
\end{equation}

\subsection{Classification}

The obtained vector representations are aggregated through mean pooling to derive a vector representation, denoted as \( \bar{z} \in \mathbb{R}^{d} \), such that:
\begin{equation} \label{equation:pooled_features}
    \bar{z} = \frac{1}{N} \sum_{i=1}^{N} z_{i} \, .
\end{equation}

$\bar{z}$ is subsequently passed to a classification head, noted $\Psi$, which predicts the logit vector $\hat{y} \in \mathbb{R}^{M}$. The model is trained on a multi-label classification task using Binary Cross-Entropy as the loss function~\citep{mao_cross-entropy_2023}.

\begin{figure*}[!ht]
	\centering
	\includegraphics[width=1.00\textwidth]{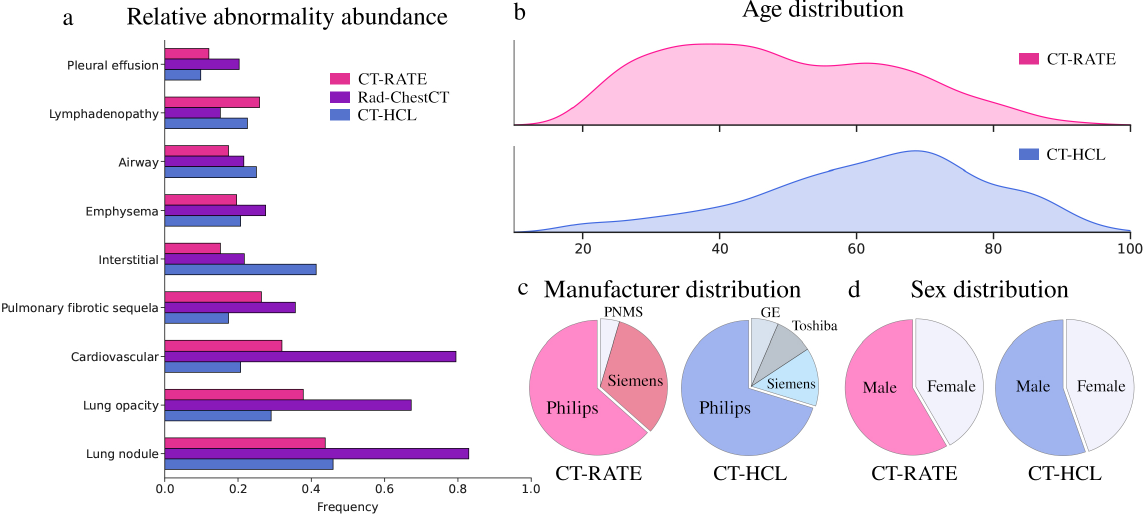}
	\caption{Comprehensive analysis of the datasets. Metadata not available for the {\tt RAD-ChestCT} dataset. a) Abnormalities from {\tt CT-HCL} are extracted with a BERT-based language model trained on french radiology reports from manually extracted anotations. b) {\tt CT-HCL} comprises data from 2,000 unique patients, with age randing from 20 to 100 years. c) {\tt CT-HCL} volumes comes from Hospices Civil de Lyon, with scanners from four manufacturers. d) {\tt CT-HCL} volumes were acquired both from male and female patients.}
    \label{fig:dataset}
\end{figure*}

%%%%%%%%%%%%%%%%%%%%%%%%%%%%%%%%%%%%%%%%%%%%%%%%%%%%%%%%%%%%%%%%%%%%%%%%%%%
% DATASET
%%%%%%%%%%%%%%%%%%%%%%%%%%%%%%%%%%%%%%%%%%%%%%%%%%%%%%%%%%%%%%%%%%%%%%%%%%%

\section{Dataset}

\paragraph{Databases.} All models are trained using 5-fold cross-validation~\citep{stone_cross-validatory_1974} and evaluated on the {\tt CT-RATE} dataset, which consists of non-contrast chest CT volumes annotated with 18 abnormalities from 21,304 unique patients~\citep{hamamci_generalist_2026}. These labels are automatically extracted from radiology reports using RadBERT~\citep{yan_radbert_2022}, a language model trained to extract abnormalities for radiology report. Duplicates are not retrained during training. To assess cross-dataset generalization, models are also evaluated on the external {\tt RAD-ChestCT} test dataset, using the 16 abnormalities shared with {\tt CT-RATE} from 1,334 unique patients, which are extracted from reports via a SARLE-based labeler~\citep{draelos_machine-learning-based_2021}. Additionally, the {\tt CT-HCL} internal dataset comprises non-contrast chest CT scans from 2,000 unique adult patients from the Hospices Civils de Lyon, with 9 abnormalities shared with {\tt CT-RATE}. These labels are manually extracted from radiology reports by radiologists~\citep{jupin-delevaux_bert-based_2023}. For cross-dataset evaluation databases ({\tt RAD-ChestCT} and {\tt CT-HCL}), exact abnormality labels do not perfectly align with those in {\tt CT-RATE}. To address this, we map related abnormalities into broader semantic groups (e.g., both \textit{Artery wall calcification} and \textit{Coronary artery wall calcification} are grouped under \textit{Calcification}). At inference time, following the protocol of the {\tt CT-RATE} original paper, the model’s prediction for each abnormality group is derived by taking the maximum predicted probability among all constituent abnormalities within that group~\citep{hamamci_generalist_2026}. This approach enables a consistent comparison across datasets despite label granularity differences. Figure~\ref{fig:dataset} provides a comprehensive comparison of the test sets from {\tt CT-RATE}, {\tt RAD-ChestCT} and {\tt CT-HCL} datasets.

\paragraph{Processing.} Consistent with prior work~\citep{hamamci_ct2rep_2024,draelos_machine-learning-based_2021,di_piazza_imitating_2025}, all datasets are processed following the same pipeline to ensure fair evaluation. CT scans are reformated to a SLP orientation, such that first axis points from inferior to superior, the second from right to left, and the third from interior to superior. Volumes are cropped or padded to a standardized resolution of $240 \times 480 \times 480$, with a spacing of 0.75 mm along the z-axis and 1.5 mm along the x- and y-axes (z, x, y). Hounsfield Units are clipped to the range [-1000, 200], which corresponds to the practical diagnostic window~\citep{denotter_hounsfield_2024}, before being mapped to the range $[0, 1]$ and normalized using ImageNet statistics (-0.449) ~\citep{russakovsky_imagenet_2015}. 

\begin{table*}[t!]
\centering

\caption{Performance of the models trained and evaluated on the {\tt CT-RATE} dataset. Mean and standard deviation are computed across 5 cross-validation folds. Experiments with (\dag) refer to a cross-dataset evaluation from models trained on {\tt CT-RATE}, and assessed on the {\tt RAD-ChestCT} and {\tt CT-HCL} datasets. \textcolor{gray!60}{Random} refer to predictions sampled from a uniform distribution. \textbf{Best} results are in bold, \underline{second-best} are underlined. {\color{RoyalBlue!75}$\blacksquare$} 3D Transformer, {\color{SkyBlue!75}$\blacksquare$} 3D CNN, {\color{Plum!75}$\blacksquare$} 2.5D.}

\begin{adjustbox}{width=1.0\textwidth}
\begin{tabular}{c l c c c c}
\toprule
\small Dataset & \small Method & \small F1-Score & \small AUROC & \small Accuracy & \small mAP\\
\toprule
% CT-RATE
\multirow{10}{*}{\rotatebox[origin=c]{0}{\small \begin{tabular}{@{}c@{}}
\small {\tt CT-RATE} \\
\end{tabular}}}& 
% random predictions
\scriptsize \color{gray!60} Random & 
$\color{gray!60} 27.78 \text{\scriptsize $\pm 0.51$}$ &
$\color{gray!60} 49.88 \text{\scriptsize $\pm 0.62$}$ &
$\color{gray!60} 49.89 \text{\scriptsize $\pm 0.31$}$ & 
$\color{gray!60} 20.94 \text{\scriptsize $\pm 0.12$}$ \\
% ViT3D
& {\color{RoyalBlue!75}$\blacksquare$} \scriptsize \textbf{ViT3D}~\citep{dosovitskiy_image_2021} & 
$49.53 \text{\textcolor{gray}{\scriptsize $\pm 0.51$}}$ & 
$78.97 \text{\textcolor{gray}{\scriptsize $\pm 0.37$}}$ & 
$75.37 \text{\textcolor{gray}{\scriptsize $\pm 0.52$}}$ & 
$46.31 \text{\textcolor{gray}{\scriptsize $\pm 0.43$}}$ \\
% ViViT
& {\color{RoyalBlue!75}$\blacksquare$} \scriptsize \textbf{ViViT}~\citep{arnab_vivit_2021} & 
$50.66 \text{\textcolor{gray}{\scriptsize $\pm 0.45$}}$ & 
$80.03 \text{\textcolor{gray}{\scriptsize $\pm 0.18$}}$ & 
$76.81 \text{\textcolor{gray}{\scriptsize $\pm 0.81$}}$ & 
$48.57 \text{\textcolor{gray}{\scriptsize $\pm 0.66$}}$ \\
% Swin3D
& {\color{RoyalBlue!75}$\blacksquare$} \scriptsize \textbf{Swin3D}~\citep{liu_video_2021} &
$50.25 \text{\textcolor{gray}{\scriptsize $\pm 0.24$}}$ & 
$79.39 \text{\textcolor{gray}{\scriptsize $\pm 0.37$}}$ & 
$76.63 \text{\textcolor{gray}{\scriptsize $\pm 0.89$}}$ & 
$47.26 \text{\textcolor{gray}{\scriptsize $\pm 0.35$}}$ \\
% ResNet3D
& {\color{SkyBlue!75}$\blacksquare$} \scriptsize \textbf{ResNet3D}~\citep{carreira_quo_2018} & 
$51.51 \text{\textcolor{gray}{\scriptsize $\pm 0.48$}}$ & 
$77.48 \text{\textcolor{gray}{\scriptsize $\pm 0.64$}}$ & 
$\underline{80.67} \text{\textcolor{gray}{\scriptsize $\pm 0.60$}}$ & 
$\underline{50.81} \text{\textcolor{gray}{\scriptsize $\pm 0.71$}}$ \\\
% CT-Net
& {\color{Plum!75}$\blacksquare$} \scriptsize  \textbf{CT-Net}~\citep{draelos_machine-learning-based_2021} &
$51.04 \text{\textcolor{gray}{\scriptsize $\pm 0.60$}}$ & 
$79.69 \text{\textcolor{gray}{\scriptsize $\pm 0.27$}}$ & 
$77.78 \text{\textcolor{gray}{\scriptsize $\pm 0.19$}}$ & 
$48.52 \text{\textcolor{gray}{\scriptsize $\pm 0.64$}}$ \\
% CT-MvG
& {\color{Plum!75}$\blacksquare$} \scriptsize \textbf{CT-MvG}~\citep{kiechle_graph_2024} & 
$52.35 \text{\textcolor{gray}{\scriptsize $\pm 0.17$}}$ & 
$81.91 \text{\textcolor{gray}{\scriptsize $\pm 0.16$}}$ & 
$78.25 \text{\textcolor{gray}{\scriptsize $\pm 0.46$}}$ & 
$51.51 \text{\textcolor{gray}{\scriptsize $\pm 0.46$}}$ \\
% CT-Scroll
& {\color{Plum!75}$\blacksquare$} \scriptsize \textbf{CT-Scroll}~\citep{di_piazza_imitating_2025} &
$\underline{54.30} \text{\textcolor{gray}{\scriptsize $\pm 0.20$}}$ & 
$\underline{82.18} \text{\textcolor{gray}{\scriptsize $\pm 0.27$}}$ & 
$79.68  \text{\textcolor{gray}{\scriptsize $\pm 0.67$}}$ & 
$53.09 \text{\textcolor{gray}{\scriptsize $\pm 0.48$}}$ \\
% CT-PSG
& \cellcolor{blue!6}{\color{Plum!75}$\blacksquare$} \scriptsize \textbf{\textcolor{RoyalBlue}{CT-SSG}} (Ours) & 
\cellcolor{blue!6}$\mathbf{57.18} \text{\textcolor{gray}{\scriptsize $\pm 0.19$}}$ &
\cellcolor{blue!6}$\mathbf{83.64} \text{\textcolor{gray}{\scriptsize $\pm 0.21$}}$ &
\cellcolor{blue!6}$\mathbf{81.03} \text{\textcolor{gray}{\scriptsize $\pm 0.38$}}$ &
\cellcolor{blue!6}$\mathbf{58.24} \text{\textcolor{gray}{\scriptsize $\pm 0.50$}}$\\

% Rad-ChestCT
\toprule
% Random predictions
\multirow{10}{*}{\rotatebox[origin=c]{0}{\small \begin{tabular}{@{}c@{}}
\small {\tt RAD-ChestCT}$^{(\dag)}$ \\
%\small \worldflag[width=0.4cm]{US}
\end{tabular}}} & 
\scriptsize \color{gray!60} Random &
$\color{gray!60} 35.91 \text{\scriptsize $\pm 0.41$}$ &
$\color{gray!60} 49.68 \text{\scriptsize $\pm 0.55$}$ &
$\color{gray!60} 50.40 \text{\scriptsize $\pm 0.32$}$ &
$\color{gray!60} 32.65 \text{\scriptsize $\pm 0.08$}$\\
% ViT3D
& {\color{RoyalBlue!75}$\blacksquare$} \scriptsize \textbf{ViT3D}~\citep{dosovitskiy_image_2021} & 
$47.94 \text{\textcolor{gray}{\scriptsize $\pm 0.85$}}$ & 
$66.85 \text{\textcolor{gray}{\scriptsize $\pm 0.18$}}$ & 
$61.42 \text{\textcolor{gray}{\scriptsize $\pm 1.90$}}$ & 
$47.94 \text{\textcolor{gray}{\scriptsize $\pm 0.32$}}$ \\
% ViViT
& {\color{RoyalBlue!75}$\blacksquare$} \scriptsize \textbf{ViViT}~\citep{arnab_vivit_2021} & 
$49.23 \text{\textcolor{gray}{\scriptsize $\pm 0.44$}}$ & 
$68.89 \text{\textcolor{gray}{\scriptsize $\pm 0.24$}}$ & 
$62.08 \text{\textcolor{gray}{\scriptsize $\pm 0.76$}}$ & 
$49.40 \text{\textcolor{gray}{\scriptsize $\pm 0.40$}}$ \\
% Swin3D
& {\color{RoyalBlue!75}$\blacksquare$} \scriptsize \textbf{Swin3D}~\citep{liu_video_2021} &
$47.36 \text{\textcolor{gray}{\scriptsize $\pm 1.10$}}$ & 
$66.34 \text{\textcolor{gray}{\scriptsize $\pm 0.45$}}$ & 
$61.34 \text{\textcolor{gray}{\scriptsize $\pm 1.41$}}$ & 
$47.71 \text{\textcolor{gray}{\scriptsize $\pm 0.66$}}$ \\
% ResNet3D
& {\color{SkyBlue!75}$\blacksquare$} \scriptsize \textbf{ResNet3D}~\citep{carreira_quo_2018} & 
$48.56 \text{\textcolor{gray}{\scriptsize $\pm 1.29$}}$ & 
$69.94 \text{\textcolor{gray}{\scriptsize $\pm 0.92$}}$ & 
$61.04 \text{\textcolor{gray}{\scriptsize $\pm 0.94$}}$ & 
$51.37 \text{\textcolor{gray}{\scriptsize $\pm 0.74$}}$ \\
% CT-Net
& {\color{Plum!75}$\blacksquare$} \scriptsize  \textbf{CT-Net}~\citep{draelos_machine-learning-based_2021} &
$49.09 \text{\textcolor{gray}{\scriptsize $\pm 0.83$}}$ & 
$69.09 \text{\textcolor{gray}{\scriptsize $\pm 0.48$}}$ & 
$62.32 \text{\textcolor{gray}{\scriptsize $\pm 0.64$}}$ & 
$49.66 \text{\textcolor{gray}{\scriptsize $\pm 0.37$}}$ \\
% CT-MvG
& {\color{Plum!75}$\blacksquare$} \scriptsize \textbf{CT-MvG}~\citep{kiechle_graph_2024} & 
$\underline{50.33} \text{\textcolor{gray}{\scriptsize $\pm 1.05$}}$ & 
$71.19 \text{\textcolor{gray}{\scriptsize $\pm 0.43$}}$ & 
$62.84 \text{\textcolor{gray}{\scriptsize $\pm 0.30$}}$ & 
$\underline{52.82} \text{\textcolor{gray}{\scriptsize $\pm 0.69$}}$ \\
% CT-Scroll
& {\color{Plum!75}$\blacksquare$} \scriptsize \textbf{CT-Scroll}~\citep{di_piazza_imitating_2025} &
$49.47 \text{\textcolor{gray}{\scriptsize $\pm 0.82$}}$ & 
$\underline{71.43} \text{\textcolor{gray}{\scriptsize $\pm 0.27$}}$ & 
$\underline{63.86} \text{\textcolor{gray}{\scriptsize $\pm 1.87$}}$ & 
$52.53 \text{\textcolor{gray}{\scriptsize $\pm 0.52$}}$ \\
% CT-PSG
& \cellcolor{blue!6}{\color{Plum!75}$\blacksquare$} \scriptsize \textbf{\textcolor{RoyalBlue}{CT-SSG}} (Ours) & 
\cellcolor{blue!6}$\mathbf{52.25} \text{\textcolor{gray}{\scriptsize $\pm 0.88$}}$ &
\cellcolor{blue!6}$\mathbf{74.58} \text{\textcolor{gray}{\scriptsize $\pm 0.36$}}$ &
\cellcolor{blue!6}$\mathbf{69.37} \text{\textcolor{gray}{\scriptsize $\pm 1.74$}}$ &
\cellcolor{blue!6}$\mathbf{58.75} \text{\textcolor{gray}{\scriptsize $\pm 0.28$}}$\\
\toprule

% CT-HCL
\multirow{10}{*}{\rotatebox[origin=c]{0}{\small \begin{tabular}{@{}c@{}}
\small {\tt CT-HCL}$^{(\dag)}$ \\
\end{tabular}}}& 
% random predictions
\scriptsize \color{gray!60} Random & 
$\color{gray!60} 33.38 \text{\scriptsize $\pm 0.28$}$ & 
$\color{gray!60} 50.16 \text{\scriptsize $\pm 0.41$}$ & 
$\color{gray!60} 49.98 \text{\scriptsize $\pm 0.40$}$ & 
$\color{gray!60} 27.01 \text{\scriptsize $\pm 0.13$}$ \\
% ViT3D
& {\color{RoyalBlue!75}$\blacksquare$} \scriptsize \textbf{ViT3D}~\citep{dosovitskiy_image_2021} & 
$44.77 \text{\textcolor{gray}{\scriptsize $\pm 0.39$}}$ & 
$64.77 \text{\textcolor{gray}{\scriptsize $\pm 0.40$}}$ & 
$52.17 \text{\textcolor{gray}{\scriptsize $\pm 0.93$}}$ & 
$41.94 \text{\textcolor{gray}{\scriptsize $\pm 0.16$}}$ \\
% ViViT
& {\color{RoyalBlue!75}$\blacksquare$} \scriptsize \textbf{ViViT}~\citep{arnab_vivit_2021} & 
$45.98 \text{\textcolor{gray}{\scriptsize $\pm 0.97$}}$ & 
$67.08 \text{\textcolor{gray}{\scriptsize $\pm 0.52$}}$ & 
$55.16 \text{\textcolor{gray}{\scriptsize $\pm 1.30$}}$ & 
$43.48 \text{\textcolor{gray}{\scriptsize $\pm 2.19$}}$ \\
% Swin3D
& {\color{RoyalBlue!75}$\blacksquare$} \scriptsize \textbf{Swin3D}~\citep{liu_video_2021} &
$44.76 \text{\textcolor{gray}{\scriptsize $\pm 0.99$}}$ & 
$64.41 \text{\textcolor{gray}{\scriptsize $\pm 0.37$}}$ & 
$52.94 \text{\textcolor{gray}{\scriptsize $\pm 0.74$}}$ & 
$41.73 \text{\textcolor{gray}{\scriptsize $\pm 0.42$}}$ \\
% ResNet3D
& {\color{SkyBlue!75}$\blacksquare$} \scriptsize \textbf{ResNet3D}~\citep{carreira_quo_2018} & 
$45.92 \text{\textcolor{gray}{\scriptsize $\pm 0.42$}}$ & 
$67.95 \text{\textcolor{gray}{\scriptsize $\pm 1.02$}}$ & 
$53.17 \text{\textcolor{gray}{\scriptsize $\pm 0.65$}}$ & 
$45.81 \text{\textcolor{gray}{\scriptsize $\pm 0.70$}}$ \\
% CT-Net
& {\color{Plum!75}$\blacksquare$} \scriptsize  \textbf{CT-Net}~\citep{draelos_machine-learning-based_2021} &
$46.81 \text{\textcolor{gray}{\scriptsize $\pm 0.83$}}$ & 
$66.53 \text{\textcolor{gray}{\scriptsize $\pm 0.49$}}$ & 
$54.70 \text{\textcolor{gray}{\scriptsize $\pm 0.33$}}$ & 
$43.16 \text{\textcolor{gray}{\scriptsize $\pm 0.43$}}$ \\
% CT-MvG
& {\color{Plum!75}$\blacksquare$} \scriptsize \textbf{CT-MvG}~\citep{kiechle_graph_2024} & 
$47.10 \text{\textcolor{gray}{\scriptsize $\pm 0.52$}}$ & 
$69.04 \text{\textcolor{gray}{\scriptsize $\pm 0.33$}}$ & 
$56.13 \text{\textcolor{gray}{\scriptsize $\pm 1.38$}}$ & 
$\underline{47.10} \text{\textcolor{gray}{\scriptsize $\pm 0.52$}}$ \\
% CT-Scroll
& {\color{Plum!75}$\blacksquare$} \scriptsize \textbf{CT-Scroll}~\citep{di_piazza_imitating_2025} &
$\underline{48.24} \text{\textcolor{gray}{\scriptsize $\pm 0.53$}}$ & 
$\underline{69.36} \text{\textcolor{gray}{\scriptsize $\pm 0.27$}}$ & 
$\underline{56.88} \text{\textcolor{gray}{\scriptsize $\pm 1.81$}}$ & 
$46.46 \text{\textcolor{gray}{\scriptsize $\pm 0.39$}}$ \\
% CT-PSG
& \cellcolor{blue!6}{\color{Plum!75}$\blacksquare$} \scriptsize \textbf{\textcolor{RoyalBlue}{CT-SSG}} (Ours) & 
\cellcolor{blue!6}$\mathbf{50.26} \text{\textcolor{gray}{\scriptsize $\pm 0.57$}}$ &
\cellcolor{blue!6}$\mathbf{71.77} \text{\textcolor{gray}{\scriptsize $\pm 0.36$}}$ &
\cellcolor{blue!6}$\mathbf{61.43} \text{\textcolor{gray}{\scriptsize $\pm 1.18$}}$ &
\cellcolor{blue!6}$\mathbf{51.81} \text{\textcolor{gray}{\scriptsize $\pm 0.44$}}$\\
\toprule

\end{tabular}
\end{adjustbox}
%\vspace{-0.8em}
\label{table:quantitive_metrics}
\end{table*}

%%%%%%%%%%%%%%%%%%%%%%%%%%%%%%%%%%%%%%%%%%%%%%%%%%%%%%%%%%%%%%%%%%%%%%%%%%%
% EXPERIMENTS
%%%%%%%%%%%%%%%%%%%%%%%%%%%%%%%%%%%%%%%%%%%%%%%%%%%%%%%%%%%%%%%%%%%%%%%%%%%

\section{Experiments}

Our experimental results comprise 4 sections: (Section~\ref{res_classification}) We provide quantitative and qualitative results on the multi-label abnormality classification task; (Section~\ref{res_ablation}) We perform an ablation study on CT-SSG components; (Section~\ref{sec:robustness}) We evaluate model's robustness to patient body translations along the z-axis and to intensity noise perturbations; (Section~\ref{res_report_generation}) We extend results to the report generation task; and (Section~\ref{res_abdominal}) we evaluate our model's transfer learning ability on CT abdominal scans for abnormality classification.

\subsection{Multi-label abnormality Classification}
\label{res_classification}

\paragraph{Baselines.} We compare our approach against three categories of baselines. First, we consider 3D Convolutional Neural Networks ({\color{SkyBlue!75}\small$\blacksquare$}) with an inflated ResNet3D~\citep{carreira_quo_2018}, which which leverages filter inflation from pretrained 2D networks to learn transferable volumetric representations. Second, we evaluate against Transformer-based architectures ({\color{RoyalBlue!75}\small$\blacksquare$}), including ViT3D, a straightforward extension of Vision Transformer to 3D volumes~\citep{dosovitskiy_image_2021}; ViViT~\citep{arnab_vivit_2021}, originally designed for video processing; and Swin3D, an adaptation of the Swin Transformer with hierarchical window-based attention for 3D inputs~\citep{liu_video_2021}. Third, we benchmark against 2.5D methods ({\color{Plum!75}\small$\blacksquare$}), which process 2D slices to extract feature maps. This includes CT-Net, which aggregates triplet axial slices features using a lightweight 3D CNN~\citep{draelos_machine-learning-based_2021}, and CT-Scroll, which employs alternating local and global attention mechanisms to capture dependencies across slices~\citep{di_piazza_imitating_2025}. Additionally, we include a Multi-View Graph (CT-MvG) baseline, which represents each 3D volume as a graph of nodes corresponding to orthogonal axial, sagittal, and coronal slices~\citep{kiechle_graph_2024}. 

To ensure a fair comparison, all models are initialized with ImageNet-pretrained weights, either directly for 2D architectures or via weight inflation~\citep{zhang_adapting_2023} for 3D counterparts, promoting stable and efficient convergence. Specifically, the ResNet3D via weight inflation~\citep{carreira_quo_2018} from a 2D ResNet-18~\citep{he_deep_2015} pretrained on ImageNet~\citep{russakovsky_imagenet_2015}; ViT3D and ViViT via weight inflation from a 2D ViT-S16~\citep{dosovitskiy_image_2021} pretrained on ImageNet; and Swin3D via weight inflation from a 2D Swin-S16~\citep{liu_swin_2021} pretrained on ImageNet. Since Vision Transformers families of models are typically release in multiple capacity variants, we adopt the \textit{Small} variants across all baselines to ensure comparability. This consistent choice offer comparable parameter counts and computational budgets, enabling a balanced comparison without favoring a particular design and representing a practically deployable setting while still retaining sufficient capacity to serve as strong baselines. Similarly, 2.5D models leveraging ResNet-18 backbones use ImageNet-pretrained 2D ResNet-18 weights at initialization. The 2D ViT-S16 module within CT-MvG is initialized with ImageNet-pretrained weights. Both for CT-SSG and baselines, all parameters are trainable during training to ensure a fair and consistent comparison across methods. Appendix~\ref{appendix:table:summary_protocol} reports the training details of all methods.

\paragraph{Evaluation protocol.} All models are trained and evaluated using 5-fold cross-validation. For each run, we apply early stopping based on the checkpoint that achieves the highest macro F1-score on the validation set, i.e. the harmonic mean of precision and recall, averaged across all abnormalities. We report performance metrics on the test set corresponding to this selected checkpoint.

\begin{figure*}[!ht]
	\centering
	\includegraphics[width=1.00\textwidth]{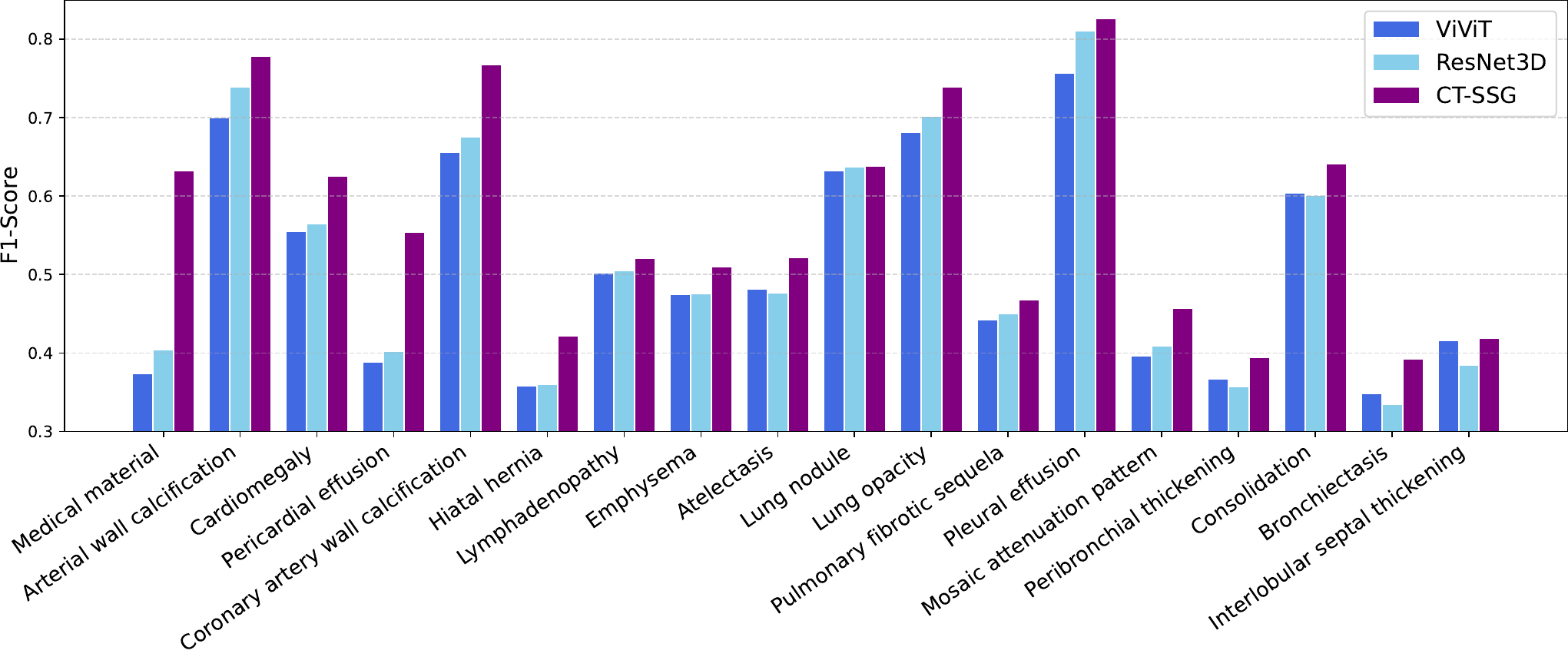}
	\caption{F1-Score per abnormality for the 18 abnormalities from the {\tt CT-RATE} test set, comparing our proposed CT-SSG with representative 3D Convolutional and 3D Transformer baselines. For clarity, one representative model per family is reported. CT-SSG consistently improves over both baselines, with the largest absolute gains observed in \textit{Pericardial effusion} (+$\Delta$8.96\%), \textit{Calcification} (+$\Delta$6.23\%), and \textit{Pleural effusion} (+$\Delta$6.20\%).}
    \label{fig:f1_per_anomaly}
\end{figure*}

\begin{figure}[t]
    \centering
    \includegraphics[width=0.92\columnwidth]{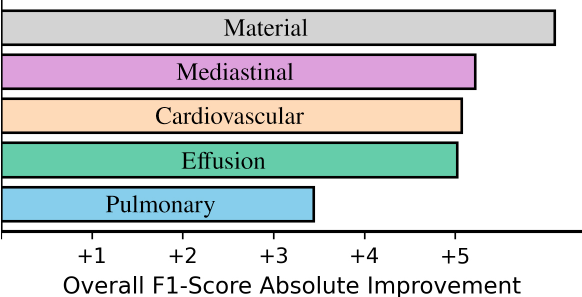}
    \caption{Average absolute F1-score improvements of CT-SSG over representative 3D convolutional (CNN) and 3D Transformer (ViViT) baselines. Abnormalities are grouped by anatomical region and pathophysiological type to highlight systematic patterns of gain. CT-SSG yields consistent improvements across groups.}
    \label{fig:f1_per_group}
\end{figure}

\paragraph{Quantitative Results.} Table~\ref{table:quantitive_metrics} reports the quantitative performance of all methods in terms of macro F1-Score, AUROC, Accuracy, and mean Average Precision (mAP). On the {\tt CT-RATE} test set, CT-SSG achieves a macro-averaged F1-Score of $57.18$, yielding relative gains of +$\Delta$5.30\% over CT-Scroll~\citep{di_piazza_imitating_2025}, +$\Delta$11.01\% over ResNet3D~\citep{anaya-isaza_overview_2021} and +$\Delta$12.87\% over ViViT~\citep{arnab_vivit_2021}. Paired t-tests across all metrics indicate $p-value < 0.01$, confirming that the improvements are statistically significant for $\alpha=0.01$, $\alpha$ being the Type I error rate~\citep{ross_paired_2017}. In cross-dataset evaluations on {\tt RAD-ChestCT} and {\tt CT-HCL}, CT-SSG consistently ranks highest across metrics, indicating strong generalization to distinct clinical distributions. Although the harmonization of the abnormality taxonomy may affect absolute scores, the relative ordering of methods is preserved across metrics and datasets.

Figure~\ref{fig:f1_per_group} reports the average absolute F1-score gain of CT-SSG over ResNet3D and ViViT, aggregated by anatomical region and pathophysiological category. Notably, CT-SSG yields a +$\Delta$3.2\% improvement for pulmonary diseases and +$\Delta$5.0\% for mediastinal, cardiovascular, and effusion diseases, indicating that the performance gains are consistent across diverse abnormality types. Figure~\ref{fig:f1_per_anomaly} presents per-abnormality F1-scores for CT-SSG, ViViT, and ResNet3D, while Table~\ref{appendix:table:metrics_labels} details the results for all baselines, showing that CT-SSG achieves superior classification performance for the majority of abnormalities. Abnormality-wise performance and t-distributed stochastic neighbor embedding (t-SNE) projections produced by CT-SSG are provided in Appendix~\ref{sec:appendix:per_abnormality_f1} and Appendix~\ref{sec:appendix:tsne}, respectively.
 
\paragraph{Contextualization with CT Foundation Models.} We additionally contextualize our approach against recent CT-specific foundation models. We include CT-CLIP pretrained with vision-language alignment from chest CT~\citep{hamamci_generalist_2026}, Merlin also pretrained with vision-language alignment from abdominal CT~\citep{blankemeier_merlin_2024}, CT-FM pretrained on CT scans with a wide variety of body parts through visual contrastive learning~\citep{pai_vision_2025}, and COLIPRI pretrained on chest CT with both vision-only, vision-language alignment, opposite sentences and report generation objectives~\citep{wald_comprehensive_2026}. As these foundation models are pretrained using objectives different from supervised abnormality classification, we evaluate them under a linear probing protocol. Specifically, the visual encoder of each pretrained model is frozen, and only a classification head is trained for multi-label abnormality prediction. Training splits and evaluation protocols are kept consistent across all methods to ensure a controlled comparison. We note, however, that the pretraining data of CT-CLIP and COLIPRI partially overlap with our validation sets used for linear probing, as these models are pretrained on the full \texttt{CT-RATE} train set. Importantly, all reported results are evaluated on strictly held-out test sets for both \texttt{CT-RATE} and \texttt{RAD-ChestCT}, with no overlap with either pretraining or linear probing data.

\begin{table}[h]
\centering
\begin{adjustbox}{width=1.00\columnwidth}
\begin{tabular}{c l c c c c}
\toprule

\multicolumn{2}{c}{Method} & \multicolumn{2}{c}{\texttt{CT-RATE}} & \multicolumn{2}{c}{\texttt{RAD-ChestCT}} \\
\cmidrule(lr){1-2} \cmidrule(lr){3-4} \cmidrule(lr){5-6}

Setup & Model & AUROC & mAP & AUROC & mAP \\
\toprule
%%%%%%%%%%%%%%%%
% Foundation
%%%%%%%%%%%%%%%%

% Model Foundation 1
\multirow{4}{*}{\rotatebox[origin=c]{0}{\small \begin{tabular}{@{}c@{}}
\small Foundation\\Probing\\
\end{tabular}}} & 
CT-FM & 
$74.35 \text{\textcolor{gray}{\scriptsize $\pm 0.09$}}$ &
$39.22 \text{\textcolor{gray}{\scriptsize $\pm 0.19$}}$ &
$63.26 \text{\textcolor{gray}{\scriptsize $\pm 0.13$}}$ &
$43.78 \text{\textcolor{gray}{\scriptsize $\pm 0.15$}}$ \\

% Foundation 2
&
CT-CLIP$^{(*)}$ & 
$76.85 \text{\textcolor{gray}{\scriptsize $\pm 0.04$}}$ &
$42.87 \text{\textcolor{gray}{\scriptsize $\pm 0.08$}}$ &
$63.38 \text{\textcolor{gray}{\scriptsize $\pm 0.05$}}$ &
$44.30 \text{\textcolor{gray}{\scriptsize $\pm 0.06$}}$ \\

% Foundation 3
&
Merlin & 
$77.85 \text{\textcolor{gray}{\scriptsize $\pm 0.04$}}$ &
$46.53 \text{\textcolor{gray}{\scriptsize $\pm 0.07$}}$ &
$69.80 \text{\textcolor{gray}{\scriptsize $\pm 0.05$}}$ &
$49.84 \text{\textcolor{gray}{\scriptsize $\pm 0.08$}}$ \\

% Foundation 4
&
COLIPRI$^{(*)}$ & 
$\mathbf{84.01} \text{\textcolor{gray}{\scriptsize $\pm 0.03$}}$ &
$\underline{57.38} \text{\textcolor{gray}{\scriptsize $\pm 0.08$}}$ &
$\underline{73.98} \text{\textcolor{gray}{\scriptsize $\pm 0.02$}}$ &
$\underline{55.45} \text{\textcolor{gray}{\scriptsize $\pm 0.10$}}$ \\

\midrule

%%%%%%%%%%%%%%%%
% Supervised
%%%%%%%%%%%%%%%%

% Model Foundation 1
Supervised & 
CT-SSG & 
$\underline{83.64} \text{\textcolor{gray}{\scriptsize $\pm 0.21$}}$ &
$\mathbf{58.24} \text{\textcolor{gray}{\scriptsize $\pm 0.50$}}$ &
$\mathbf{74.58} \text{\textcolor{gray}{\scriptsize $\pm 0.36$}}$ &
$\mathbf{58.75} \text{\textcolor{gray}{\scriptsize $\pm 0.28$}}$  \\

\bottomrule
\end{tabular}
\end{adjustbox}
%\vspace{-0.8em}
\caption{
Comparison of linear probing on frozen visual encoders from foundation models with CT-SSG, on \texttt{CT-RATE}. \textbf{Best results} in bold, \underline{second-best} underlined. $^{(*)}$Foundation models are pretrained on data that partially overlap with ours \texttt{CT-RATE} validation splits. However, all reported evaluations are conducted on strictly held-out test sets, with no overlap for either \texttt{CT-RATE} or \texttt{RAD-ChestCT}.
}
\label{table:quantitative_foundation}
\end{table}

Table~\ref{table:quantitative_foundation} shows that CT-SSG outperforms CT-FM, CT-CLIP and Merlin on \texttt{CT-RATE}, in terms of both AUROC and mAP, achieving the best mAP among all compared methods. While COLIPRI attains the highest AUROC on \texttt{CT-RATE}, CT-SSG achieves a closely comparable performance. On \texttt{RAD-ChestCT}, CT-SSG yields the best results in both AUROC and mAP when compared to CT-specific foundation models (paired t-test across folds, $p<0.01$). These results indicate that CT-SSG's learned representations are well suited for multi-label abnormality analysis and generalize well to unseen datasets, even when compared with representations derived from foundation models with substantially higher architectural complexity trained using generic large-scale pretraining objectives. Additional details on the evaluated CT foundation models, including pretraining objectives, data processing, model complexity and feature extraction protocols, are provided in Appendix~\ref{sec:appendix:foundation}.

\begin{figure*}[!ht]
	\centering
	\includegraphics[width=1.00\textwidth]{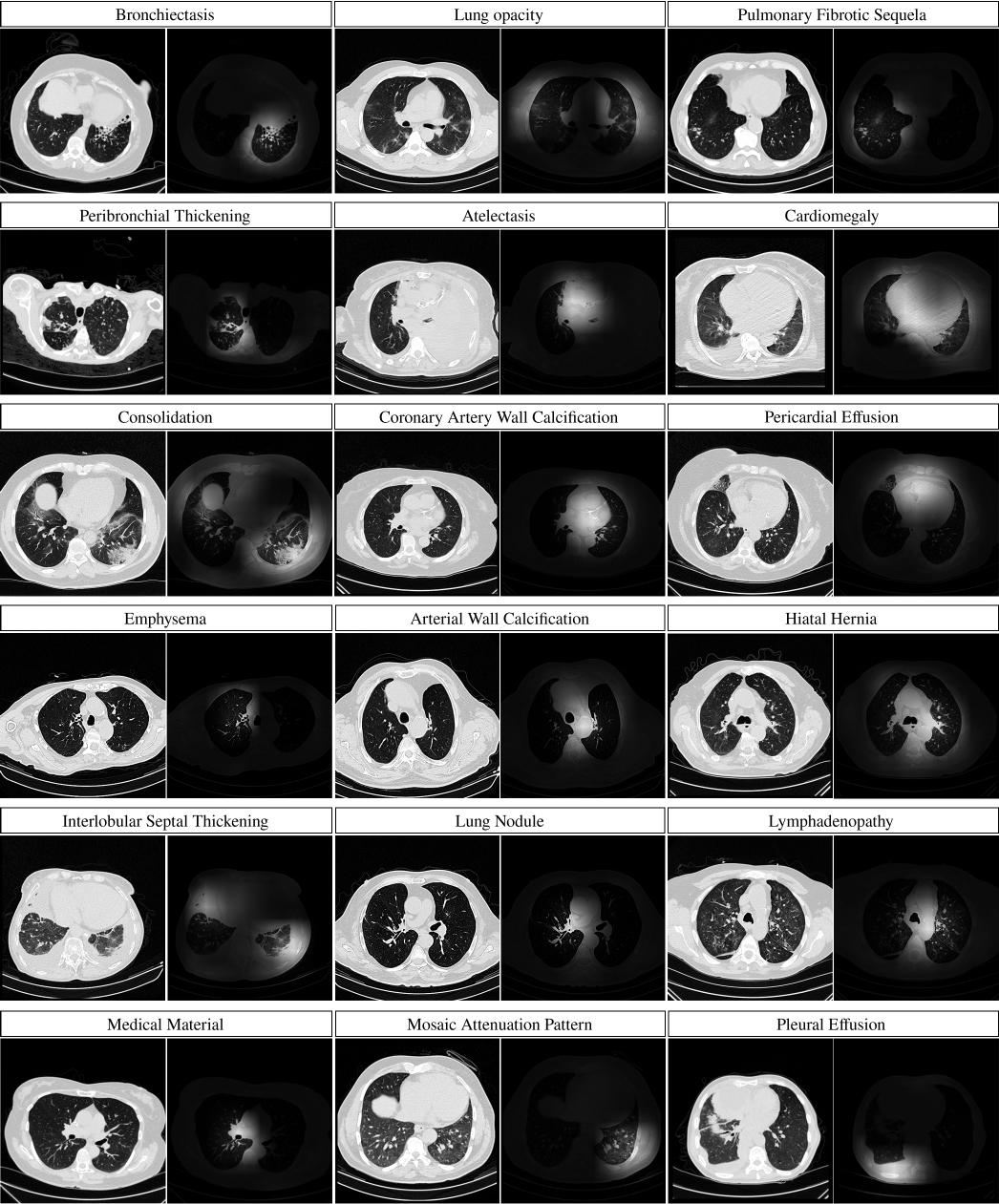}
	\caption{Gradient-weighted class activation maps, extracted from the 2D ResNet from the triplet slices embeddings module, where darker regions indicate lower activations. For each input, we display the slice with the highest absolute activation value from the heatmap.}
    \label{fig:gradcam}
\end{figure*}

\paragraph{Qualitative Results.} In addition to the quantitative analysis, Figure~\ref{fig:gradcam} presents qualitative examples of correct predictions for each abnormality in the {\tt CT-RATE} dataset. We visualize Gradient-Weight Class Activation Mapping~\citep{selvaraju_grad-cam_2019} heatmaps, where darker regions correspond to lower activations. Specifically for each volume, we obtain a heatmap of shape $S \times H_{s} \times W_{s}$ and we display the $s$-th axial slice with highest activation, highlighting CT-SSG's ability to classify abnormalities from relevant regions.

\subsection{Ablation Studies}
\label{res_ablation}

\begin{table}[h]
\centering
\begin{adjustbox}{width=1.00\columnwidth}
\begin{tabular}{l c c c c}
\toprule
Component & AUROC & F1-Score & $\Delta$F1 (abs) & $\Delta$F1 (\%)\\
\toprule
% Feature extraction
Triplet slices embeddings & 
$81.20 \text{\textcolor{gray}{\scriptsize $\pm 0.42$}}$ &
$52.41 \text{\textcolor{gray}{\scriptsize $\pm 0.69$}}$ &
N/A &
N/A \\
+ Spatial convolution & 
$82.45 \text{\textcolor{gray}{\scriptsize $\pm 0.19$}}$ &
$54.25 \text{\textcolor{gray}{\scriptsize $\pm 0.37$}}$ &
$+1.84$ &
$+3.51$ \\
% Spectral block
+ Spectral convolution & 
$83.06 \text{\textcolor{gray}{\scriptsize $\pm 0.09$}}$ &
$55.43 \text{\textcolor{gray}{\scriptsize $\pm 0.44$}}$ &
$+1.18$ &
$+2.18$ \\
% Layer Normalization
+ Normalization layer & 
$83.28 \text{\textcolor{gray}{\scriptsize $\pm 0.19$}}$ &
$56.16 \text{\textcolor{gray}{\scriptsize $\pm 0.39$}}$ &
$+0.73$ &
$+1.32$ \\
% Residual Connections
+ Residual connection & 
$83.32 \text{\textcolor{gray}{\scriptsize $\pm 0.16$}}$ &
$56.37 \text{\textcolor{gray}{\scriptsize $\pm 0.13$}}$ &
$+0.21$ &
$+0.37$ \\
% With edge weighting
+ Edge weighting & 
$83.37 \text{\textcolor{gray}{\scriptsize $\pm 0.33$}}$ &
$56.54 \text{\textcolor{gray}{\scriptsize $\pm 0.15$}}$ &
$+0.17$ &
$+0.30$ \\
% Positional encoding
+ Axial. pos. encoding & 
$83.53 \text{\textcolor{gray}{\scriptsize $\pm 0.23$}}$ &
$56.76 \text{\textcolor{gray}{\scriptsize $\pm 0.31$}}$ &
$+0.22$ &
$+0.39$ \\
% Sparse connectivity
+ Sparse topology & 
$83.64 \text{\textcolor{gray}{\scriptsize $\pm 0.21$}}$ &
$57.18 \text{\textcolor{gray}{\scriptsize $\pm 0.19$}}$ &
$+0.42$ &
$+0.74$ \\
\toprule
\end{tabular}
\end{adjustbox}
%\vspace{-0.8em}
\caption{Incremental contribution of each model component, on the {\tt CT-RATE} test set. Starting from the base architecture, components are added cumulatively across rows. For each step, we report F1-score and AUROC, along with absolute and relative improvements ($\Delta$F1) over the configuration in the preceding row. All components yield consistent gains, indicating that the overall performance arises from complementary contributions.}
\label{table:quantitative_ablation_study}
\end{table}

\paragraph{Impact of each model component.} Table~\ref{table:quantitative_ablation_study} quantifies the incremental impact of each component. Starting from the initialization of the node features, we progressively integrate seven architectural elements. Each addition results in a positive improvement in both F1-score and AUROC. For example, the integration of spectral network yields a +$\Delta$2.18\% improvement in F1-Score over spatial convolution, while the axial positional encoding module results in a +$\Delta$0.39\% improvement. The cumulative trend underscores that the model benefits from the synergistic effect of multiple design choices, which suggests that each component addresses complementary aspects of the task and collectively ensures robustness.

\paragraph{Impact of node design.} To investigate the impact of slice selection, Table~\ref{table:node_design} compare a graph construction with triplets of adjacent slices to one with triplets formed by repeating the same slice three times. Using adjacent axial slices for the three input channels yields consistently better performance than repeating a single slice across channels, suggesting that incorporating limited inter-slice context is beneficial for abnormality classification.

\begin{table}[h]
\centering
\begin{adjustbox}{width=1.00\columnwidth}
\begin{tabular}{l l c c c}
\toprule
Node design & Slice indices & F1-Score & AUROC & mAP\\
\toprule

% Adjacent
Adjacent slices & 
$(s, s+1, s+2)$ &
$\underline{57.18} \text{\textcolor{gray}{\scriptsize $\pm 0.19$}}$ &
$\underline{83.64} \text{\textcolor{gray}{\scriptsize $\pm 0.21$}}$ &
$\underline{58.24} \text{\textcolor{gray}{\scriptsize $\pm 0.50$}}$ \\

% Same
Single-slice (x3) & 
$(s, s, s)$ &
$56.64 \text{\textcolor{gray}{\scriptsize $\pm 0.26$}}$ &
$83.34 \text{\textcolor{gray}{\scriptsize $\pm 0.29$}}$ &
$57.31 \text{\textcolor{gray}{\scriptsize $\pm 0.25$}}$  \\

\toprule
\end{tabular}
\end{adjustbox}
%\vspace{-0.8em}
\caption{
Comparison between adjacent-slice and single-slice triplets replicated across channels. Adjacent triplets consistently outperform replicated single-slice representations, highlighting the benefit of local inter-slice context. \underline{Best results} are underlined.
}
\label{table:node_design}
\end{table}

\begin{figure}[h]
    \centering
    \includegraphics[width=\columnwidth]{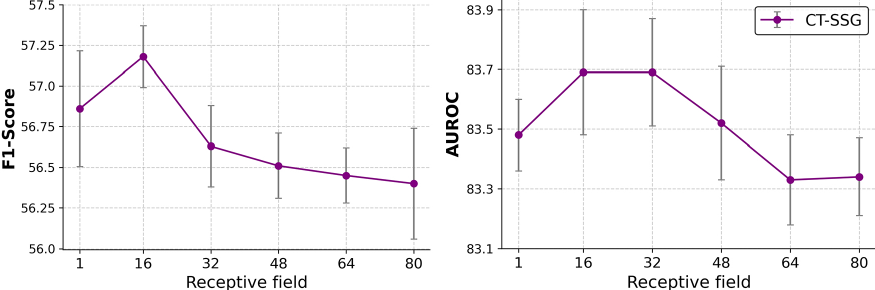}
    \caption{Impact of the receptive field. We compare different receptive field sizes in $\{1, 16, 32, 64, 80\}$, where $80$ is the maximum number of nodes. Higher F1-Score and AUROC is observed when restricting the aggregation to a local neighborhood.}
    \label{fig:receptive_field}
\end{figure}

\paragraph{Impact of receptive field.} We further analyze the impact of the receptive field, defined as the number of neighboring nodes connected to each node, corresponding to the number of axial slice triplets considered for feature aggregation. Figure~\ref{fig:receptive_field} reports F1-Score and AUROC for receptive field sizes in $\{1, 16, 32, 64, 80\}$, where 80 corresponds to the maximum number of triplets in a CT scan. Higher performance is observed for smaller receptive fields, with peak F1-Score at 16 neighbors in our settings, suggesting that restricting the aggregation to a local neighborhood facilitates effective integration of relevant visual context for abnormality classification.

\paragraph{Impact of model depth.} Table~\ref{table:impact_depth} details multi-label abnormality classification performances with 1, 3, and 5 spectral blocks. Increasing the number of propagation layers does not improve performance. In fact, a single layer yields the best results. This indicates that most discriminative information resides in immediate slice-to-slice dependencies, and that effective modeling of 3D CT does not necessarily require deep graph structures, but rather careful design of local inter-slice connectivity.

\begin{table}[h]
\centering
\begin{adjustbox}{width=1.00\columnwidth}
\begin{tabular}{c c c c c}
\toprule
Model depth $L$ & F1-Score & AUROC & mAP & Accuracy\\
\toprule
% L=1
$1$ & 
$\underline{57.18} \text{\textcolor{gray}{\scriptsize $\pm 0.19$}}$ &
$\underline{83.64} \text{\textcolor{gray}{\scriptsize $\pm 0.21$}}$ &
$\underline{58.24} \text{\textcolor{gray}{\scriptsize $\pm 0.50$}}$ &
$81.03 \text{\textcolor{gray}{\scriptsize $\pm 0.38$}}$ \\
% L=3
$3$ & 
$56.46 \text{\textcolor{gray}{\scriptsize $\pm 0.20$}}$ &
$83.04 \text{\textcolor{gray}{\scriptsize $\pm 0.09$}}$ &
$56.58 \text{\textcolor{gray}{\scriptsize $\pm 0.35$}}$ &
$81.44 \text{\textcolor{gray}{\scriptsize $\pm 0.60$}}$ \\
% L=5
$5$ & 
$56.47 \text{\textcolor{gray}{\scriptsize $\pm 0.16$}}$ &
$83.09 \text{\textcolor{gray}{\scriptsize $\pm 0.06$}}$ &
$56.79 \text{\textcolor{gray}{\scriptsize $\pm 0.25$}}$ &
$\underline{82.11} \text{\textcolor{gray}{\scriptsize $\pm 0.51$}}$ \\
\toprule
\end{tabular}
\end{adjustbox}
%\vspace{-0.8em}
\caption{Impact of the model depth. We compare models with 1, 3, and 5 layers. Performance peaks at a single layer, suggesting that shallow inter-slice message passing is sufficient for effective representation learning.}
\label{table:impact_depth}
\end{table}

\paragraph{Impact of the spectral filter size.} Spectral polynomial filters present \guillemotleft \ two sources of locality \guillemotright. First, the adjacency matrix defines \textit{who is a neighbor of whom}. For a fully connected graph, every node would be a 1-\textit{hop connected to every other node}. Second, the spectral filter size $K$ which defines \textit{how far the Laplacian power is applied}. For a fully connected graph, every node is already directly 1-hop away from every other node, which means that increasing $K$ does not expand the receptive field but just adds higher-order polynomials of the Laplacian. However if the graph is sparse, the spectral filter size $K$ truly increases the receptive field, making the spectral filter exactly $K$-localized. Hence, we systematically evaluate the impact of spectral filter size $K$, which controls the Chebyshev polynomial order in Equation~\ref{equation:cheb}, under different graph topology: fully connected and sparse. 

Table~\ref{table:impact_filter_size} summarizes the effect of varying the spectral filter size $K \in {1,3,5}$ across different graph topologies. On the fully connected topology, the best results yields for $K=5$, but the improvement is not statistically significant at the conventional $\alpha=0.05$ (paired t-test, $p=0.077$), though it approaches significance. In contrast, on the sparse graph with receptive field fixed to $16$, a larger filter size proves beneficial: $K=3$ achieves the highest F1-score ($57.18$), and the improvement over $K=1$ is statistically significant ($p<0.01$). These findings highlight the dual role of connectivity and spectral filter size in shaping the effective receptive field of the network. In fully connected graphs, even small Chebyshev orders ($K=1$) suffice to capture global context, since each node already aggregates information from all others. Increasing $K$ in this setting may lead to redundant propagation and potential over-smoothing, which explains why larger filters do not improve performance. In contrast, under constrained receptive fields (e.g., sparse graphs with fixed neighborhood size), higher-order filters become crucial. A moderate filter size ($K=3$) expands the receptive field sufficiently to integrate useful multi-hop context while avoiding the noise introduced by excessively large filters. This suggests that spectral filters primarily act as a mechanism to compensate for sparsity in connectivity, while in dense regimes their effect diminishes or even harms performance.

\begin{table}[h]
\centering
\begin{adjustbox}{width=1.00\columnwidth}
\begin{tabular}{c c c c c}
\toprule
Topology & Filter size $K$ & F1-Score & AUROC & mAP\\
\toprule

% Topology: Fully-connected
\multirow{3}{*}{\rotatebox[origin=c]{0}{\small \begin{tabular}{@{}c@{}}
\small Fully connected\\$q=80$
\end{tabular}}}& 
% K=1
$1$ & 
$56.76 \text{\textcolor{gray}{\scriptsize $\pm 0.41$}}$ &
$83.63 \text{\textcolor{gray}{\scriptsize $\pm 0.23$}}$ &
$57.56 \text{\textcolor{gray}{\scriptsize $\pm 0.47$}}$ \\

% K=3
& $3$ & 
$56.40\text{\textcolor{gray}{\scriptsize $\pm 0.44$}}$ &
$83.34 \text{\textcolor{gray}{\scriptsize $\pm 0.13$}}$ &
$57.39 \text{\textcolor{gray}{\scriptsize $\pm 0.31$}}$ \\

% K=5
& $5$ & 
$56.83 \text{\textcolor{gray}{\scriptsize $\pm 0.43$}}$ &
$83.56 \text{\textcolor{gray}{\scriptsize $\pm 0.19$}}$ &
$57.69 \text{\textcolor{gray}{\scriptsize $\pm 0.32$}}$ \\

\toprule

% Topology: Sparse
\multirow{3}{*}{\rotatebox[origin=c]{0}{\small \begin{tabular}{@{}c@{}}
\small Sparse\\$q=16$
\end{tabular}}}& 
% K=1
$1$ & 
$56.50 \text{\textcolor{gray}{\scriptsize $\pm 0.19$}}$ &
$83.63 \text{\textcolor{gray}{\scriptsize $\pm 0.21$}}$ &
$57.57 \text{\textcolor{gray}{\scriptsize $\pm 0.21$}}$ \\
% K=3
& $3$ & 
$\underline{57.18} \text{\textcolor{gray}{\scriptsize $\pm 0.19$}}$ &
$\underline{83.64} \text{\textcolor{gray}{\scriptsize $\pm 0.21$}}$ &
$\underline{58.24} \text{\textcolor{gray}{\scriptsize $\pm 0.50$}}$ \\
% K=5
& $5$ & 
$56.41 \text{\textcolor{gray}{\scriptsize $\pm 0.31$}}$ &
$83.37 \text{\textcolor{gray}{\scriptsize $\pm 0.19$}}$ &
$57.35 \text{\textcolor{gray}{\scriptsize $\pm 0.22$}}$ \\

\toprule
\end{tabular}
\end{adjustbox}
%\vspace{-0.8em}
\caption{Impact of the spectral filter size $K$, for different graph topologies. $q$ refers to the receptive field.}
\label{table:impact_filter_size}
\end{table}

\paragraph{Impact of convolutional operator.} Table~\ref{table:ablation_modules} summarizes abnormality classification performance across different graph operators. Among them, the Chebyshev spectral convolution~\citep{defferrard_convolutional_2017} achieves the strongest results, both on fully connected and sparse topologies. In particular, the spectral model consistently outperforms its spatial counterparts, yielding a relative improvement of +$\Delta$2.82\% in F1-Score compared to Graph Attention~\citep{velickovic_graph_2018}, and +$\Delta$0.65\% compared to Graph Convolution~\citep{morris_weisfeiler_2021}. These results highlight the advantage of leveraging spectral formulations to capture dependencies across slices, and suggest that Graph Convolution may be too restrictive, while Graph Attention that can be more expressive, often require more parameters and larger training size to be competitive. 

\paragraph{Impact of graph topology.}
Building on this operator-level analysis, we next examine how graph connectivity influences performance. Specifically, we compare a fully connected topology, analogous to the Transformer formulation where all nodes (triplet of axial slices) attend to one another, with a sparse topology that constrains interactions to local neighborhoods along the caudal–cranial axis. Across all operators, Table~\ref{table:ablation_modules} shows that sparse graphs consistently outperform fully connected ones. On average, sparse topologies yield a +$\Delta$0.82\% improvement in F1-Score, suggesting that limiting the receptive field to local interactions better captures short-range dependencies between adjacent slices and ultimately enhances abnormality classification performance.

\begin{table}[h]
\centering
\begin{adjustbox}{width=1.00\columnwidth}
\begin{tabular}{l l l c c}
\toprule
Network & Operator & Topology & F1-Score & AUROC\\
\toprule
\multirow{4}{*}{\rotatebox[origin=c]{0}{\small \begin{tabular}{@{}c@{}}
\small Spatial \\
\end{tabular}}} 

% GraphConv
& \multirow{2}{*}{\rotatebox[origin=c]{0}{\small \begin{tabular}{@{}c@{}}
\small Graph Conv. \\
\end{tabular}}}& 
\small Fully connected &
$56.35 \text{\textcolor{gray}{\scriptsize $\pm 0.18$}}$ &
$83.12 \text{\textcolor{gray}{\scriptsize $\pm 0.22$}}$ \\
% GraphConv - [80, 32, 16]
& & 
\small Sparse &
$\underline{56.81} \text{\textcolor{gray}{\scriptsize $\pm 0.28$}}$ &
$\underline{83.44} \text{\textcolor{gray}{\scriptsize $\pm 0.25$}}$ \\

%\toprule
\cmidrule{2-5}
% GATv2
& \multirow{2}{*}{\rotatebox[origin=c]{0}{\small \begin{tabular}{@{}c@{}}
\small Graph Attention \\
\end{tabular}}} & 
\small Fully connected &
$55.05 \text{\textcolor{gray}{\scriptsize $\pm 0.19$}}$ &
$82.81 \text{\textcolor{gray}{\scriptsize $\pm 0.07$}}$ \\
% GraphConv - [80, 32, 16]
& & 
\small Sparse &
$\underline{55.61} \text{\textcolor{gray}{\scriptsize $\pm 0.16$}}$ &
$\underline{82.98} \text{\textcolor{gray}{\scriptsize $\pm 0.18$}}$ \\

\toprule
\multirow{2}{*}{\rotatebox[origin=c]{0}{\small \begin{tabular}{@{}c@{}}
\small Spectral \\
\end{tabular}}}

% ChebConv
& \multirow{2}{*}{\rotatebox[origin=c]{0}{\small \begin{tabular}{@{}c@{}}
\small Chebyshev \\
\end{tabular}}} &
% ChebConv - Fully connected
\small Fully connected &
$56.83 \text{\textcolor{gray}{\scriptsize $\pm 0.43$}}$ &
$83.56 \text{\textcolor{gray}{\scriptsize $\pm 0.19$}}$ \\
% ChebConv - Sparse
& & 
\small Sparse &
$\underline{57.18} \text{\textcolor{gray}{\scriptsize $\pm 0.19$}}$ &
$\underline{83.64} \text{\textcolor{gray}{\scriptsize $\pm 0.21$}}$ \\

\toprule
\end{tabular}
\end{adjustbox}
%\vspace{-0.8em}
\caption{Impact of graph topology on different convolutional operators. Results are reported on the classification task on the {\tt CT-RATE} test set. The sparse topology is defined with receptive field size $q=16$. Graph Convolution and Graph Attention operators are implemented with \href{https://pytorch-geometric.readthedocs.io/en/latest/generated/torch_geometric.nn.conv.GraphConv.html}{GraphConv} and \href{https://pytorch-geometric.readthedocs.io/en/latest/generated/torch_geometric.nn.conv.GATv2Conv.html}{GATv2} from \href{https://pytorch-geometric.readthedocs.io/en/latest/modules/nn.html}{PyTorch Geometrics}. For each operator, \underline{best results} are underlined.}
\label{table:ablation_modules}
\end{table}

\subsection{Robustness Analysis}
\label{sec:robustness}

Beyond overall classification accuracy, a clinically deployable model must remain reliable under common sources of variation in CT acquisitions. To this end, we assess the robustness of our approach to two perturbation types: \textit{z-axis translation}, simulating patient body translation along the caudal–cranial axis~\citep{di_piazza_structured_2025}, and \textit{robustness to noise}, simulating variations in scanner calibration or patient-specific attenuation~\citep{kiechle_graph_2024}. Given the large number of baseline methods, we report robustness comparisons against the strongest representative of each family, as identified in our quantitative evaluations. This strategy eases understability and facilitates clearer interpretation.

\begin{figure}[h]
    \centering
    \includegraphics[width=\columnwidth]{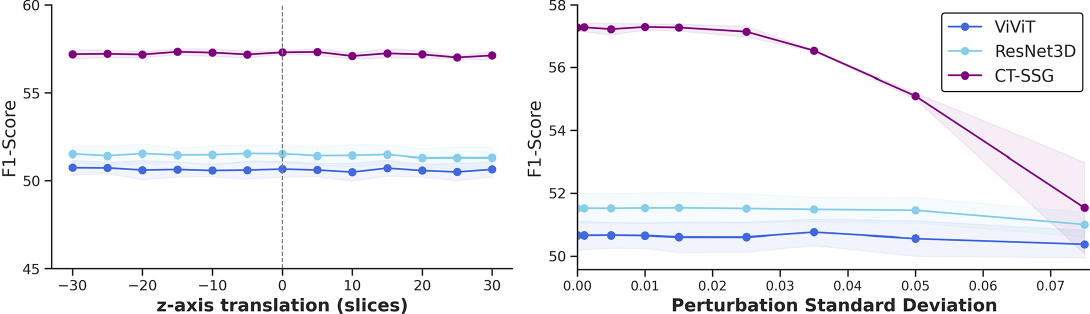}
    \caption{Robustness evaluation. \textit{Left}: macro-F1 under axial $z$-axis translations ($-30$ to $+30$ slices), where all methods remain invariant to volumetric shifts. \textit{Right}: macro-F1 under Gaussian noise perturbations of increasing standard deviation, where performance is stable up to $\sigma=0.025$ and CT-SSG maintains higher F1-Score than baselines even as noise increases.}
    \label{fig:robustness_analysis}
\end{figure}

\paragraph{Sensitivity to patient body translation.} To emulate variability in patient positioning along the caudal–cranial axis, we translate the input volume by $5$ to $30$ axial slices in both directions, applying minimum-value padding to preserve dimensional consistency. Across all methods, Figure~\ref{fig:robustness_analysis} shows that macro-F1 remains unchanged, indicating that both CT-SSG and the 3D-modeling baselines are robust to moderate volumetric misalignments.

\paragraph{Sensitivity to noise.} Following prior work~\citep{sudre_longitudinal_2017,kiechle_graph_2024}, we simulate acquisition-related variations by injecting Gaussian noise into voxel intensities, with standard deviation ranging from $\sigma_{min} = 0.01$ corresponding to low noise, to $\sigma_{max} = 0.07$ corresponding to high noise. While CT-SSG exhibits a mild decrease in performance at higher noise levels, it consistently maintains an advantage over the supervised baseline. This pattern suggests that pretraining confers robustness to common intensity variations.

\subsection{Transfer on the Automated Report Generation task}
\label{res_report_generation}

\begin{figure*}[h]
	\centering
	\includegraphics[width=1.00\textwidth]{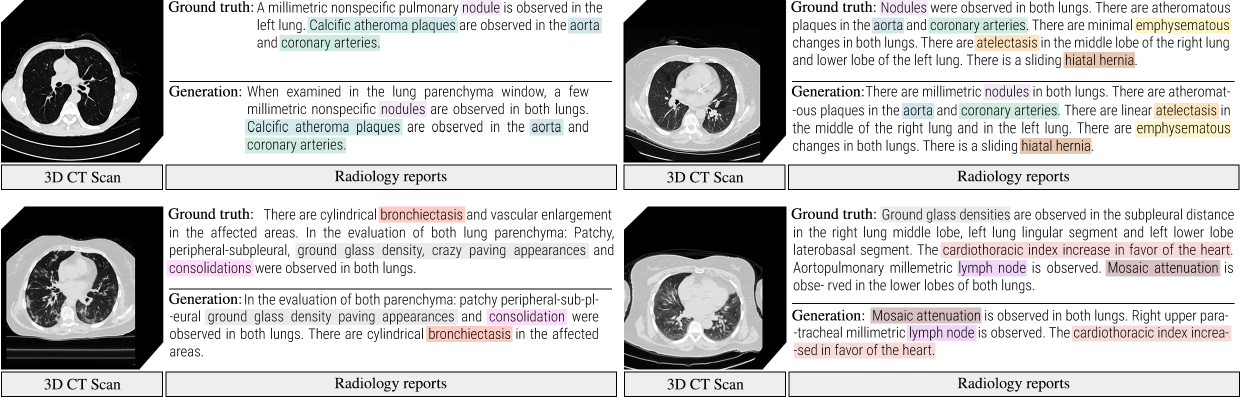}
	\caption{Qualitative comparison between ground-truth reports and those generated with CT-SSG for 3D Chest CT volumes. Color-coded highlights indicate abnormalities correctly captured by the model, demonstrating alignment with ground-truth annotations.}
    \label{fig:report_generation_qualitative}
\end{figure*}

\paragraph{Motivation.} We further evaluate the visual encoders on an automated report generation task. To ensure that differences in downstream performance reflect the quality of the learned visual representations rather than decoder engineering, we adopt a deliberately simple encoder-decoder architecture inspired by CT2Rep~\citep{hamamci_ct2rep_2024}. Concretely, the visual encoder is pretrained on the multi-label abnormality classification task and kept frozen while a lightweight decoder is trained with a next-token prediction objective. This setup isolates the quality of the latent space, which is the central focus of our study, from the confounding effects or more sophisticated sequence modeling strategies. Integration of advanced components such as large pretrained language models~\citep{li_2_2025}, extensive modality-specific pretraining~\citep{blankemeier_merlin_2024} or multimodal fusion~\citep{liu_enhanced_2025} if left for future work.

\begin{figure}[h]
    \centering
    \includegraphics[width=\columnwidth]{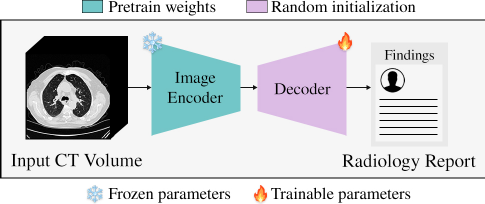}
    \caption{Report generation framework overview. The frozen pretrained image encoder extract visual features that are given to a decoder which generates the report, in a auto-regressively manner.}
    \label{fig:encoder_decoder}
\end{figure}

\paragraph{Evaluation protocol.} Figure~\ref{fig:encoder_decoder} provides an overview of the encoder-decoder pipeline. The pretrained visual encoder is frozen and a lightweight decoder is trained using a token-level cross-entropy loss for next-token prediction~\citep{karpathy_deep_2015}. At inference, reports are generated auto-regressively~\citep{vinyals_show_2015}. This protocol ensures that observed differences in generated reports are primarily attributable to differences in the visual latent representations.

\paragraph{Quantitative results.} We evaluate models using both Natural Language Generation (NLG) metrics and Clinical Efficacy (CE) metrics. NLG metrics, including BLEU-1~\citep{papineni_bleu_2002} and METEOR~\citep{lavie_meteor_2009}, assess the semantic alignment between ground-truth and generated reports. To assess clinical relevance, CE metrics quantify the model’s ability to accurately identify and report pathologies present in the 3D CT Scans. Specifically, generated reports are given to a RadBERT~\citep{yan_radbert_2022} to extract predicted abnormalities as binary label vectors. These are compared against ground-truth annotations, enabling computation of standard classification metrics. We report F1-score and also incorporate the CRG Score, a recently proposed metric that weights positive predictions based on class frequency, reflecting the clinical imperative of minimizing false negatives, which can have serious consequences in medical diagnosis~\citep{hamamci_crg_2025}. Table~\ref{table:quant_report_generation} shows that our proposed CT-SSG achieves substantial improvements over baseline encoders on the report generation task. CT-SSG yields a relative gain of +$\Delta$40.56\% in F1-Score compared to ViViT, and +$\Delta$52.14\% compared to CNN-based baseline. A paired t-test between CT-SSG and each baseline results in $p$-values below 0.01 across all metrics, confirming the statistical significance of these improvements. Finally, Table~\ref{appendix:table:report_generation_details} reports all metrics.

\begin{table}[h]
\centering
\begin{adjustbox}{width=1.00\columnwidth}
\begin{tabular}{l c c c c}
\toprule
& \multicolumn{2}{c}{NLG} & \multicolumn{2}{c}{CE}\\
\cmidrule(lr){2-3} \cmidrule(lr){4-5}

Encoder & BLEU-1 & METEOR & F1-Score & CRG\\
\toprule
% ViVIT
{\color{RoyalBlue}$\blacksquare$} ViViT & 
$0.290 \text{\textcolor{gray}{\scriptsize $\pm 0.008$}}$ &
$0.150 \text{\textcolor{gray}{\scriptsize $\pm 0.004$}}$ &
$27.54 \text{\textcolor{gray}{\scriptsize $\pm 0.40$}}$ &
$40.01 \text{\textcolor{gray}{\scriptsize $\pm 0.22$}}$ \\
% CNN
{\color{SkyBlue}$\blacksquare$} ResNet3D & 
$0.286 \text{\textcolor{gray}{\scriptsize $\pm 0.012$}}$ &
$0.148 \text{\textcolor{gray}{\scriptsize $\pm 0.004$}}$ &
$25.45 \text{\textcolor{gray}{\scriptsize $\pm 0.90$}}$ &
$39.87 \text{\textcolor{gray}{\scriptsize $\pm 0.27$}}$ \\
% CT-SSG
{\color{Plum}$\blacksquare$} CT-SSG & 
$\underline{0.317} \text{\textcolor{gray}{\scriptsize $\pm 0.001$}}$ &
$\underline{0.164} \text{\textcolor{gray}{\scriptsize $\pm 0.003$}}$ &
$\underline{38.72} \text{\textcolor{gray}{\scriptsize $\pm 0.25$}}$ &
$\underline{43.66} \text{\textcolor{gray}{\scriptsize $\pm 0.15$}}$ \\
\toprule
\end{tabular}
\end{adjustbox}
%\vspace{-0.8em}
\caption{Quantitative evaluation on the report generation task, reporting both Natural Language Generation (NLG) and Clinical Efficacy (CE) metrics on the {\tt CT-RATE} dataset. CT-SSG achieves consistent improvements over baselines across metrics, with the \underline{best results} underlined, demonstrating its ability to capture structured 3D information that benefits downstream report generation.}
\label{table:quant_report_generation}
\end{table}

\paragraph{Qualitative results.} Figure~\ref{fig:report_generation_qualitative} presents qualitative examples comparing ground-truths with the generated reports. CT-SSG consistently identifies clinically relevant abnormalities and employs appropriate medical terminology, producing outputs that closely resemble radiologist-authored reports. Both quantitative and qualitative evaluations confirm that the model captures the presence of key abnormalities. However, descriptions of spatial localization and severity remain less reliable. Although automated report generation is not the primary focus of this study, these results underscore the robustness of the learned representations and suggest a promising avenue for future research in connecting abnormality representation learning with clinically faithful language generation.

\subsection{Transfer to Abdominal CT Scans for Multi-label Abnormality Classification}
\label{res_abdominal}

\begin{figure}[h]
    \centering
    \includegraphics[width=\columnwidth]{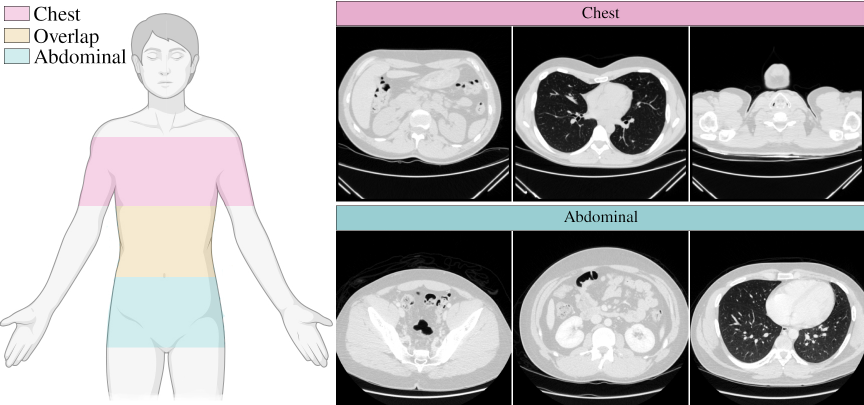}
    \caption{Comparison of chest and abdominal CT volumes. Top row: representative axial slices (first, center, last) from a chest CT volume, spanning from caudal to cranial directions (left to right). Bottom row: corresponding slices from an abdominal CT volume processed with the same spatial dimensions.}
    \label{fig:chest_abdominal}
\end{figure}

\paragraph{Motivation.} We further investigate the \textit{cross-domain generalization} of our learned representations by extending evaluation to a distinct anatomical region: 3D abdominal CT scans. Figure~\ref{fig:chest_abdominal} suggests that this setting presents a substantial domain shift, as abdominal CTs capture a different anatomical field of view and spatial context compared to chest CTs.

\paragraph{Evaluation protocol.} To assess cross-anatomy generalization, we evaluate CT-SSG chest-pretrained model on the {\tt Merlin Abdominal CT} dataset~\citep{blankemeier_merlin_2024}. We focus on the 17 labels shared with {\tt CT-RATE}, effectively probing whether features learned from chest CTs can transfer to detect the same abnormalities in a partially overlapping abdominal field of view. Expert annotations are not available, so we derive binary pseudo-labels from radiology reports using large language model-based inference~\citep{reichenpfader_scoping_2025}. While these labels may introduce some noise, this approach provides a scalable way to test whether chest-pretrained representations encode medical priors that generalize across anatomical domains and imaging contexts.

\begin{figure}[h]
    \centering
    \includegraphics[width=\columnwidth]{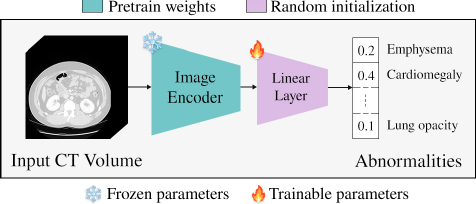}
    \caption{Linear probe framework overview. The frozen pretrained image encoder extract visual features that are given to a linear layer to predict abnormalities.}
    \label{fig:linear_probe}
\end{figure}

Following established evaluation transfer protocols~\citep{misra_self-supervised_2019,bardes_vicreg_2021}, we compare two configurations: a \textit{supervised baseline} in which CT-SSG is trained from scratch with ImageNet-initialized ResNet weights, and a \textit{linear probing}, in which a linear classifier is trained on top of frozen representations from our chest-pretrained CT-SSG backbone. Figure~\ref{fig:linear_probe} illustrates the Linear probe framework. We vary the training set size between $250$ and $10,000$ samples, with validation and test sets each comprising $1,000$ unique patients, ensuring subject-level separation.

\begin{figure}[h]
    \centering
    \includegraphics[width=\columnwidth]{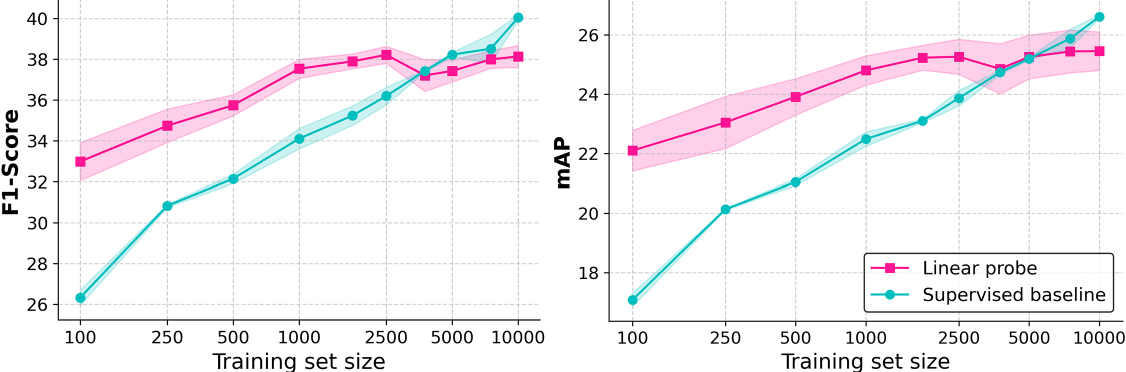}
    \caption{Transfer to abdominal CT. Comparison of a linear probe trained on frozen chest-pretrained CT-SSG representations against a supervised CT-SSG trained from scratch on the {\tt Merlin Abdominal CT} dataset~\citep{blankemeier_merlin_2024}. Performance is reported in terms of macro-F1 and macro-mAP for different size of the train set.}
    \label{fig:merlin_f1_vs_trainsizes}
\end{figure}

\paragraph{Quantitative results.} Figure~\ref{fig:merlin_f1_vs_trainsizes} reports macro-F1 and macro-mAP across training set sizes. In the low-data regime (100-3,750 samples), the \textit{linear probe} consistently outperforms the supervised baseline, demonstrating that chest-pretrained representations encode transferable structural and textural priors that enable sample-efficient adaptation to abdominal CT. At larger training sizes, the fully supervised baseline surpasses the linear probe, as the increased availability of labeled data allows the model to specialize to the abdominal domain. These findings underscore the practical value of pretraining for clinically realistic, label-scarce scenarios.

\subsection{Implementation Details}
\label{section:implementation_details}

\paragraph{Multi-label Abnormality Classification.} CT-SSG was trained using the Adam optimizer~\citep{kingma_adam_2017} with $(\beta_{1}, \beta_{2})=(0.9, 0.99)$ and a learning rate of $0.0001$. We used a batch size of $4$, with $10,000$ warm-up steps and $200,000$ iterations to ensure convergence. The training duration of CT-SSG was one day on a single NVIDIA RTX A6000 GPU. We used a GPU with $48$GB of memory but the training can only require 16GB of memory with implementation of gradient accumulation. Inference takes approximately 70 milliseconds.

\paragraph{Report Generation.} The Encoder-Decoder report generation framework from Section~\ref{res_report_generation}, was trained using Adam with a batch size of $4$, $(\beta_{1}, \beta_{2})=(0.9, 0.99)$, and a learning rate of $0.00005$ for $400,000$ iterations to ensure convergence. At inference, we use the Beam Search algorithm as generation mode~\citep{freitag_beam_2017} with a beam size set to $4$, and generated sentences were limited to $300$ tokens. Inference takes approximately $0.90$ seconds per generated report.

\paragraph{Linear probe.} The linear probe on abdominal CTs, presented in Section~\ref{res_abdominal}, was trained using Adam with $(\beta_{1}, \beta_{2})=(0.9, 0.99)$, a learning rate of $0.0001$, and a batch size of $4$ for $200,000$ iterations.

%%%%%%%%%%%%%%%%%%%%%%%%%%%%%%%%%%%%%%%%%%%%%%%%%%%%%%%%%%%%%%%%%%%%%%%%%%%
% CONCLUSION
%%%%%%%%%%%%%%%%%%%%%%%%%%%%%%%%%%%%%%%%%%%%%%%%%%%%%%%%%%%%%%%%%%%%%%%%%%%
\section{Conclusion and Discussion}
In this work, we introduced CT-SSG, a 2.5D approach that represents 3D CT Volumes as structured graphs constructed from axial slices. Evaluated on chest CT datasets for multi-label abnormality classification, CT-SSG achieves competitive performance while maintaining computational efficiency compatible with clinical deployment. 
Specifically, restricting inter-slice connectivity to local neighborhoods, rather than adopting a fully-connected transformer-style topology, yields higher clinical accuracy, suggesting that explicit structural priors can serve as an effective inductive bias for 3D reasoning. Our ablation studies confirm that graph topology, positional encoding, and aggregation operators play complementary roles in enabling this sparse yet expressive representation.
Beyond classification, we demonstrated the transferability of the learned representations to radiology report generation and abdominal CT, highlighting the robustness and generality of the proposed approach.

\paragraph{Limitations and Future work.}

(1) Our work primarily focuses on graph-based feature aggregation and employs a 2D ResNet-18 encoder for slice-level feature initialization. While this choice allows us to isolate the contribution of the proposed aggregation mechanism, the CT-SSG formulation remains encoder-agnostic. Future research could investigate the integration of higher-capacity visual encoders and advanced pretraining strategies, which may further improve representation learning and overall performance. (2) Since CT-SSG relies on spectral convolutions, the learned filters are linked to the underlying graph topology. Although the proposed graph construction yields highly consistent structures across volumes due to spatial standardization and anatomical ordering, substantial variations in acquisition protocols, field-of-view, or graph construction strategies may require adaptation or retraining. Future work could investigate topology-robust graph learning or domain adaptation strategies to improve generalization across heterogeneous imaging settings. (3) As a 2.5D, axial-slice–based method, CT-SSG does not fully exploit volumetric information. A promising direction is to hybridize with a fully 3D branch or adopt multi-view representations that integrate sagittal and coronal planes to complement axial features. (4) In transferring to automated report generation, CT-SSG consistently identifies key abnormalities but is less reliable in describing their spatial localization and severity. As report generation was not the primary objective of this study, these findings underscore the versatility of the learned representations while pointing toward an important future direction: bridging abnormality representation learning with clinically faithful narrative generation.

%%%%%%%%%%%%%%%%%%%%%%%%%%%%%%%%%%%%%%%%%%%%%%%%%%%%%%%%%%%%%%%%%%%%%%%
% Mandatory Sections. Please complete, especially for final publication
%%%%%%%%%%%%%%%%%%%%%%%%%%%%%%%%%%%%%%%%%%%%%%%%%%%%%%%%%%%%%%%%%%%%%%%

% Acknowledgements.
% Please include any funding, intellectual contributions not included in the authorship, and any other acknowledgements.
\acks{We acknowledge {\tt CT-RATE}, {\tt RAD-ChestCT}, and {\tt Merlin CT} authors for releasing their public datasets to be used for this work of academic research. The patient body icon from Figure~\ref{fig:chest_abdominal} was created with \textit{BioRender.com}. Finally, we thank the anonymous reviewers and editors of the MICCAI EMERGE Workshop and of the MELBA Journal for their valuable feedback and suggestions.}

% Ethical Standards.
% Please edit with the appropriate ethics considerations for your work. Include any pertinent IRB information, etc.
%
% Please note that the submission requirements included:
% The work presented must follow appropriate ethical standards in conducting research and writing the manuscript, following all applicable laws and regulations regarding treatment of animals or human subjects.
\ethics{The work follows appropriate ethical standards in conducting research and writing the manuscript, following all applicable laws and regulations regarding treatment of animals or human subjects.}

% Conflict of Interest
% Declaration of possible conflicts of interest: Authors must disclose any financial, organisational, commercial or personal conflicts of interest that might bias their work.
% If no conflicts, please say "We declare we don't have conflicts of interest."
\coi{The authors declare no conflict of interest.}

% Data availability
\data{The {\tt CT-RATE} public dataset is available at \url{https://huggingface.co/datasets/ibrahimhamamci/CT-RATE}. The {\tt RAD-ChestCT} public dataset is available at \url{https://zenodo.org/records/6406114}. The {\tt Merlin Abdominal CT} public dataset is available at \url{https://stanfordaimi.azurewebsites.net/datasets/60b9c7ff-877b-48ce-96c3-0194c8205c40}.}

\bibliography{reference}

% needed in general before appendices.
% \newpage

% Appendix is optional
\clearpage
%\onecolumn
\appendix

\FloatBarrier

\section{Per-abnormality F1-Score} \label{sec:appendix:per_abnormality_f1}

Table~\ref{appendix:table:metrics_labels} presents a per-abnormality performance comparison between CT-SSG and competing baseline methods. All models are trained and evaluated on the {\tt CT-RATE} dataset, and the reported results correspond to the mean performance across a 5-fold cross-validation.

%%%%%%%%%%%%%%%%%%%%%%%%%%%%%%%%%%%%%%%%%%%%%
% Per-abnormality F1-Score with all baselines
%%%%%%%%%%%%%%%%%%%%%%%%%%%%%%%%%%%%%%%%%%%%%
\renewcommand{\arraystretch}{1.5}
\begin{table*}[h]
\centering
\small
\begin{adjustbox}{width=1.0\textwidth}
\begin{tabular}{p{3.75cm} *{8}{>{\centering\arraybackslash}p{1.75cm}}}
\hline 
\textbf{Abnormality} & \textbf{ViT3D} & \textbf{ViViT} & \textbf{Swin3D} & \textbf{ResNet3D} & \textbf{CT-Net} & \textbf{CT-MvG} & \textbf{CT-Scroll} & \textbf{CT-SSG} \\
\hline
Medical material & 
0.374 & % ViT3D
0.373 & % ViViT
0.380 & % Swin3D
0.403 & % ResNet3D
0.450 & % CT-Net
0.393 & % CT-MvG
0.540 & % CT-Scroll
\underline{0.631} \\ % SSG
\rowcolor[gray]{0.9}
Arterial wall calcification &
0.701 & % ViT3D
0.699 & % ViViT
0.692 & % Swin3D
0.738 & % ResNet3D
0.771 & % CT-Net
0.716 & % CT-MvG
0.758 & % CT-Scroll
\underline{0.778} \\ % SSG
Cardiomegaly &
0.551 & % ViT3D
0.554 & % ViViT
0.608 & % Swin3D
0.564 & % ResNet3D
0.586 & % CT-Net
0.574 & % CT-MvG
0.587 & % CT-Scroll
\underline{0.624} \\ % SSG
\rowcolor[gray]{0.9}
Pericardial effusion &
0.401 & % ViT3D
0.388 & % ViViT
0.400 & % Swin3D
0.401 & % ResNet3D
0.385 & % CT-Net
0.378 & % CT-MvG
0.420 & % CT-Scroll
\underline{0.554} \\ % SSG
Coronary artery wall calcif. &
0.647 & % ViT3D
0.655 & % ViViT
0.628 & % Swin3D
0.674 & % ResNet3D
0.760 & % CT-Net
0.656 & % CT-MvG
0.768 & % CT-Scroll
0.767 \\ % SSG
\rowcolor[gray]{0.9}
Hiatal hernia &
0.360 & % ViT3D
0.357 & % ViViT
0.358 & % Swin3D
0.359 & % ResNet3D
0.346 & % CT-Net
0.365 & % CT-MvG
0.362 & % CT-Scroll
\underline{0.421} \\ % SSG
Lymphadenopathy &
0.504 & % ViT3D
0.502 & % ViViT
0.499 & % Swin3D
0.504 & % ResNet3D
0.487 & % CT-Net
0.522 & % CT-MvG
0.511 & % CT-Scroll
0.520 \\ % SSG
\rowcolor[gray]{0.9}
Emphysema &
0.467 & % ViT3D
0.474 & % ViViT
0.470 & % Swin3D
0.475 & % ResNet3D
0.450 & % CT-Net
0.506 & % CT-MvG
0.487 & % CT-Scroll
\underline{0.509} \\ % SSG
Atelectasis &
0.454 & % ViT3D
0.481 & % ViViT
0.469 & % Swin3D
0.476 & % ResNet3D
0.466 & % CT-Net
0.503 & % CT-MvG
0.494 & % CT-Scroll
\underline{0.521} \\ % SSG
\rowcolor[gray]{0.9}
Lung nodule &
0.632 & % ViT3D
0.632 & % ViViT
0.633 & % Swin3D
0.637 & % ResNet3D
0.633 & % CT-Net
0.637 & % CT-MvG
0.634 & % CT-Scroll
\underline{0.637} \\ % SSG
Lung opacity &
0.638 & % ViT3D
0.681 & % ViViT
0.676 & % Swin3D
0.701 & % ResNet3D
0.716 & % CT-Net
0.703 & % CT-MvG
0.738 & % CT-Scroll
\underline{0.738} \\ % SSG
\rowcolor[gray]{0.9}
Pulmonary fibrotic sequela &
0.441 & % ViT3D
0.442 & % ViViT
0.443 & % Swin3D
0.449 & % ResNet3D
0.433 & % CT-Net
0.457 & % CT-MvG
0.456 & % CT-Scroll
\underline{0.467} \\ % SSG
Pleural effusion &
0.768 & % ViT3D
0.756 & % ViViT
0.776 & % Swin3D
0.810 & % ResNet3D
0.776 & % CT-Net
0.778 & % CT-MvG
0.817 & % CT-Scroll
\underline{0.825} \\ % SSG
\rowcolor[gray]{0.9}
Mosaic attenuation &
0.361 & % ViT3D
0.395 & % ViViT
0.387 & % Swin3D
0.408 & % ResNet3D
0.338 & % CT-Net
0.423 & % CT-MvG
0.394 & % CT-Scroll
\underline{0.456} \\ % SSG
Peribronchial thickening &
0.355 & % ViT3D
0.366 & % ViViT
0.343 & % Swin3D
0.356 & % ResNet3D
0.343 & % CT-Net
0.394 & % CT-MvG
0.384 & % CT-Scroll
\underline{0.394} \\ % SSG
\rowcolor[gray]{0.9}
Consolidation &
0.558 & % ViT3D
0.603 & % ViViT
0.604 & % Swin3D
0.600 & % ResNet3D
0.560 & % CT-Net
0.625 & % CT-MvG
0.641 & % CT-Scroll
\underline{0.641} \\ % SSG
Bronchiectasis &
0.292 & % ViT3D
0.347 & % ViViT
0.304 & % Swin3D
0.333 & % ResNet3D
0.304 & % CT-Net
0.373 & % CT-MvG
0.367 & % CT-Scroll
\underline{0.391} \\ % SSG
\rowcolor[gray]{0.9}
Interlobular septal thick. &
0.413 & % ViT3D
0.415 & % ViViT
0.377 & % Swin3D
0.384 & % ResNet3D
0.374 & % CT-Net
0.418 & % CT-MvG
0.417 & % CT-Scroll
\underline{0.418} \\
\hline
\textbf{Mean} & 
0.495 & % ViT3D
0.507 & % ViViT
0.503 & % Swin3D
0.515 & % ResNet3D
0.510 & % CT-Net
0.524 & % CT-MvG
0.543 & % CT-Scroll
\underline{0.572} \\ % SSG
\hline
\end{tabular}
\end{adjustbox}
\caption{Per-abnormality F1-Score of all methods evaluated on the {\tt CT-RATE} test set, averaged across 5 folds. The \underline{underlined metrics} are those that have improved with CT-SSG compared to baselines. \textit{Calcif.} denotes for \textit{calcification} and \textit{thick.} for \textit{thickening}. CT-SSG demonstrates improved F1-Scores across the majority of abnormalities.}
\label{appendix:table:metrics_labels}
\end{table*}

%%%%%%%%%%%%%%%%%%%%%%%%%%%%%%%%%%%%%%%%%%%%%
% Report Generation detailed metrics
%%%%%%%%%%%%%%%%%%%%%%%%%%%%%%%%%%%%%%%%%%%%%
\renewcommand{\arraystretch}{1.5}
\begin{table*}[h]
\centering
\small
\begin{adjustbox}{width=1.0\textwidth}
\begin{tabular}{p{2.00cm} *{8}{>{\centering\arraybackslash}p{1.5cm}}}
\hline 
& \multicolumn{4}{c}{Natural Language Generation} & \multicolumn{4}{c}{Clinical Efficacy} \\
\cmidrule(lr){2-5} \cmidrule(lr){6-9}
\textbf{Encoder} & \textbf{BLEU-1} & \textbf{BLEU-4} & \textbf{METEOR} & \textbf{ROUGE} & \textbf{F1-Score} & \textbf{Precision} & \textbf{Recall} & \textbf{Accuracy} \\
\hline

ResNet3D & 
0.286 & % BLEU1
0.106 & % BLEU4
0.148 & % METEOR
0.242 & % ROUGE
0.254 & % F1
0.384 & % Precision
0.236 & % Recall
0.816 \\ % Accuracy

ViViT & 
0.290 & % BLEU1
0.106 & % BLEU4
0.150 & % METEOR
0.241 & % ROUGE
0.275 & % F1
0.418 & % Precision
0.253 & % Recall
0.808 \\ % Accuracy

\cellcolor[gray]{0.925}CT-SSG & 
\cellcolor[gray]{0.925}0.317 & % BLEU1
\cellcolor[gray]{0.925}0.107 & % BLEU4
\cellcolor[gray]{0.925}0.164 & % METEOR
\cellcolor[gray]{0.925}0.246 & % ROUGE
\cellcolor[gray]{0.925}0.387 & % F1
\cellcolor[gray]{0.925}0.520 & % Precision
\cellcolor[gray]{0.925}0.363 & % Recall
\cellcolor[gray]{0.925}0.814 \\ % Accuracy

\hline
\end{tabular}
\end{adjustbox}
\caption{Detailed automated report generation metrics evaluated on the {\tt CT-RATE} test set, averaged across 5 folds.}
\label{appendix:table:report_generation_details}
\end{table*}

%%%%%%%%%%%%%%%%%%%%%%%%%%%%%%%%%%%%%%%%%%%%%
% Per-abnormality detailed metrics for CT-SSG
%%%%%%%%%%%%%%%%%%%%%%%%%%%%%%%%%%%%%%%%%%%%%
\renewcommand{\arraystretch}{1.5}
\begin{table*}[h]
\centering
\small
\begin{adjustbox}{width=1.0\textwidth}
%\begin{tabular}{l c c c c c c}
\begin{tabular}{p{4.75cm} *{6}{>{\centering\arraybackslash}p{2.00cm}}}
\hline 
\textbf{Abnormality} & \textbf{F1-Score} & \textbf{Precision} & \textbf{Recall} & \textbf{AUROC} & \textbf{AP} & \textbf{Accuracy} \\
\hline
Medical material & 
0.631 & % macro-F1
0.561 & % macro-P
0.674 & % macro-R
0.905 & % AUROC
0.647 & % AP
0.914\\ % Accuracy
\rowcolor[gray]{0.9}
Arterial wall calcification &
0.778 & % macro-F1
0.729 & % macro-P
0.814 & % macro-R
0.928 & % AUROC
0.812 & % AP
0.865\\ % Accuracy
Cardiomegaly &
0.624 & % macro-F1
0.532 & % macro-P
0.702 & % macro-R
0.935 & % AUROC
0.660 & % AP
0.901\\ % Accuracy
\rowcolor[gray]{0.9}
Pericardial effusion &
0.554 & % macro-F1
0.499 & % macro-P
0.625 & % macro-R
0.902 & % AUROC
0.586 & % AP
0.928\\ % Accuracy
Coronary artery wall calcification &
0.767 & % macro-F1
0.680 & % macro-P
0.866 & % macro-R
0.931 & % AUROC
0.782 & % AP
0.869\\ % Accuracy
\rowcolor[gray]{0.9}
Hiatal hernia &
0.421 & % macro-F1
0.322 & % macro-P
0.563 & % macro-R
0.773 & % AUROC
0.410 & % AP
0.770\\ % Accuracy
Lymphadenopathy &
0.520 & % macro-F1
0.419 & % macro-P
0.693 & % macro-R
0.743 & % AUROC
0.509 & % AP
0.676\\ % Accuracy
\rowcolor[gray]{0.9}
Emphysema &
0.509 & % macro-F1
0.474 & % macro-P
0.518 & % macro-R
0.789 & % AUROC
0.546 & % AP
0.794\\ % Accuracy
Atelectasis &
0.521 & % macro-F1
0.441 & % macro-P
0.675 & % macro-R
0.776 & % AUROC
0.543 & % AP
0.726\\ % Accuracy
\rowcolor[gray]{0.9}
Lung nodule &
0.637 & % macro-F1
0.493 & % macro-P
0.907 & % macro-R
0.626 & % AUROC
0.540 & % AP
0.547\\ % Accuracy
Lung opacity &
0.738 & % macro-F1
0.735 & % macro-P
0.742 & % macro-R
0.866 & % AUROC
0.824 & % AP
0.801\\ % Accuracy
\rowcolor[gray]{0.9}
Pulmonary fibrotic sequela &
0.467 & % macro-F1
0.350 & % macro-P
0.722 & % macro-R
0.683 & % AUROC
0.462 & % AP
0.567\\ % Accuracy
Pleural effusion &
0.825 & % macro-F1
0.826 & % macro-P
0.832 & % macro-R
0.969 & % AUROC
0.853 & % AP
0.959\\ % Accuracy
\rowcolor[gray]{0.9}
Mosaic attenuation pattern &
0.456 & % macro-F1
0.498 & % macro-P
0.410 & % macro-R
0.884 & % AUROC
0.488 & % AP
0.919\\ % Accuracy
Peribronchial thickening &
0.394 & % macro-F1
0.324 & % macro-P
0.554 & % macro-R
0.798 & % AUROC
0.399 & % AP
0.808\\ % Accuracy
\rowcolor[gray]{0.9}
Consolidation &
0.641 & % macro-F1
0.575 & % macro-P
0.721 & % macro-R
0.901 & % AUROC
0.662 & % AP
0.846\\ % Accuracy
Bronchiectasis &
0.391 & % macro-F1
0.327 & % macro-P
0.504 & % macro-R
0.787 & % AUROC
0.422 & % AP
0.838\\ % Accuracy
\rowcolor[gray]{0.9}
Interlobular septal thicking &
0.418 & % macro-F1
0.369 & % macro-P
0.446 & % macro-R
0.860 & % AUROC
0.365 & % AP
0.892\\ % Accuracy
\hline
\textbf{Mean} & 
0.572 & % macro-F1
0.509 & % macro-P
0.665 & % macro-R
0.836 & % AUROC
0.582 & % AP
0.812\\ % Accuracy
\hline
\end{tabular}
\end{adjustbox}
\caption{Per-abnormality classification metrics for CT-SSG, evaluated on the \texttt{CT-RATE} test set and averaged across 5 folds. We report macro F1-Score with Precision, Recall, AUROC, Average Precision (AP) and Accuracy.}
\label{appendix:table:metrics_detailed}
\end{table*}

% --- Capacity-Performance ---
\section{Model capacity and performance} \label{sec:appendix:capacity_performance}

\begin{figure*}[h]
    \centering
    \includegraphics[width=0.95\textwidth]{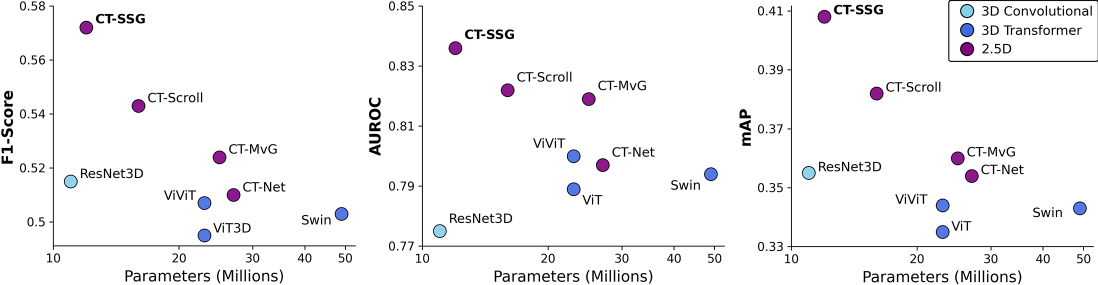}
    \caption{Comparison of F1-Score, AUROC and mAP across models with varying parameter counts. Despite comparable model capacities, CT-SSG achieves consistently higher performance, suggesting that the performance gains stem from improved representation learning rather than increased model size.}
    \label{fig:parameters}
\end{figure*}

To verify that the observed performance improvements are not merely due to increased model capacity, Figure~\ref{fig:parameters} displays the F1-Score, AUROC and mAP against the number of learnable parameters. Our method achieves superior performance with a comparable parameter count, indicating that the gains arise from the proposed spectral representation rather than model scaling.

% --- Pseudo-Code ---
\section{Pseudo code} \label{sec:appendix:per_anomaly_f1}

% --- PSEUDO CODE TABLE ---
\renewcommand{\arraystretch}{1.5}
\begin{table*}[h]
\centering
\begin{adjustbox}{width=1.0\textwidth}
\begin{tabular}{c l l c c c}
\hline
\textbf{Section} & \textbf{Pseudo code} & \textbf{Description} & \textbf{Input shape} & \textbf{Output shape} & \textbf{Python module}\\
\hline

%%% SECTION FEATURES INIT
\multirow{6}{*}{\rotatebox[origin=c]{0}{\small \begin{tabular}{@{}c@{}}
\small Features\\Initialization
\end{tabular}}}& 

% Load volume
$\mathcal{X} \leftarrow \text{LoadCTVolume}()$ & 
Load 3D CT Volumes &
-- &
$(B, 1, S, H_{s}, W_{s})$ &
-- \\

% Reshape
& 
\cellcolor[gray]{0.925} $\mathcal{X}^{\text{triplet}} \leftarrow \text{MakeTriplet}(\mathcal{X})$ &
\cellcolor[gray]{0.925}Group triplet of axial slices &
\cellcolor[gray]{0.925}$(B, 1, S, H_{s}, W_{s})$ &
\cellcolor[gray]{0.925}$(B, N, 3, H_{s}, W_{s})$ &
\cellcolor[gray]{0.925}\texttt{torch.reshape}$^{1}$ \\

% ResNet
& $H^{\text{resnet}} \leftarrow \text{ResNet}(\mathcal{X}^{\text{triplet}})$ & 
2D ResNet forward pass &
$(B, N, 3, H_{s}, W_{s})$ &
$(B, N, d, h, w)$ &
\texttt{resnet18}$^{2}$ \\

% GAP
& 
\cellcolor[gray]{0.925}$\bar{H} \leftarrow \text{GAP}(H^{\text{resnet}})$ & 
\cellcolor[gray]{0.925}Global Average Pooling &
\cellcolor[gray]{0.925}$(B, N, d, h, w)$ &
\cellcolor[gray]{0.925}$(B, N, d)$ &
\cellcolor[gray]{0.925}\texttt{torch.mean}$^{1}$ \\

% Positional Embedding
& $H \leftarrow \bar{H} + P_{\text{pos}}^{\text{axial}}$ & 
Positional Embedding &
$(B, N, d)$ &
$(B, N, d)$ &
-- \\

% Node feature init
&
\cellcolor[gray]{0.925}$H_{0} \leftarrow H$ & 
\cellcolor[gray]{0.925}Features initialization &
\cellcolor[gray]{0.925}$(B, N, d)$ &
\cellcolor[gray]{0.925}$(B, N, d)$ &
\cellcolor[gray]{0.925}-- \\

\hline
%%% SECTION GNN
\multirow{9}{*}{\rotatebox[origin=c]{0}{\small \begin{tabular}{@{}c@{}}
\small Message\\Passing\\($\times L$)
\end{tabular}}}& 

% LayerNorm1
${H}^{'}_{l} \leftarrow \text{LayerNorm}(H_{l})$ & 
Layer Normalization &
$(B, N, d)$ &
$(B, N, d)$ &
\texttt{nn.LayerNorm}$^{1}$ \\

% LoadEdgesIndex
& 
\cellcolor[gray]{0.925}$E^{\text{index}} \leftarrow \text{LoadEdgeIndex}()$ & 
\cellcolor[gray]{0.925}Load edges index &
\cellcolor[gray]{0.925}-- &
\cellcolor[gray]{0.925}$(2, N \times N)$ &
\cellcolor[gray]{0.925}\texttt{torch.Tensor}$^{1}$ \\

% LoadEdgesWeight
&
$E^{\text{weight}} \leftarrow \text{LoadEdgeWeight}()$ & 
Load edges weight &
-- &
($N \times N$) &
\texttt{torch.Tensor}$^{1}$ \\

% BuildGraph
& 
\cellcolor[gray]{0.925}$\mathcal{G} \leftarrow \text{BuildGraph}({H}^{'}_{l}, E^{\text{index}}, E^{\text{weight}})$ & 
\cellcolor[gray]{0.925}Build graph &
\cellcolor[gray]{0.925}-- &
\cellcolor[gray]{0.925}-- &
\cellcolor[gray]{0.925}\texttt{pyg.data.Data}$^{3}$ \\

% Chebyshev Convolution
&
${H}^{'}_{l} \leftarrow \text{ChebConv}(\mathcal{G})$ & 
Chebyshev Convolution &
$(B, N, d)$ &
$(B, N, d)$ &
\texttt{pyg.nn.conv.Chebconv}$^{3}$ \\

% Residual Connection 1
&
\cellcolor[gray]{0.925}${Z}_{l} \leftarrow H^{'}_{l} + H_{l}$ & 
\cellcolor[gray]{0.925}Residual Connection &
\cellcolor[gray]{0.925}$(B, N, d)$ &
\cellcolor[gray]{0.925}$(B, N, d)$ &
\cellcolor[gray]{0.925}-- \\

% LayerNorm2
&
${Z}^{'}_{l} \leftarrow \text{LayerNorm}({Z}_{l})$ & 
Layer Normalization &
$(B, N, d)$ &
$(B, N, d)$ &
\texttt{nn.LayerNorm}$^{1}$ \\

% FNN
&
\cellcolor[gray]{0.925}${Z}^{'}_{l} \leftarrow \text{FFN}({Z}^{'}_{l})$ & 
\cellcolor[gray]{0.925}Feedforward Neural Network &
\cellcolor[gray]{0.925}$(B, N, d)$ &
\cellcolor[gray]{0.925}$(B, N, d)$ &
\cellcolor[gray]{0.925}\texttt{nn.Linear, nn.GELU}$^{1}$ \\

% Residual Connection 2
&
${H}_{l+1} \leftarrow {Z}^{'}_{l} + Z_{l}$ & 
Residual Connection &
$(B, N, d)$ &
$(B, N, d)$ &
-- \\

\hline
%%% SECTION POOLING
\multirow{2}{*}{\rotatebox[origin=c]{0}{\small \begin{tabular}{@{}c@{}}
\small Classification
\end{tabular}}}& 

% Pooling
\cellcolor[gray]{0.925}$\bar{z} \leftarrow \text{Pooling}({H}_{L})$ & 
\cellcolor[gray]{0.925}Graph pooling &
\cellcolor[gray]{0.925}$(B, N, d)$ &
\cellcolor[gray]{0.925}$(B, d)$ &
\cellcolor[gray]{0.925}\texttt{torch.mean}$^{1}$ \\

% Classifier
&
$\hat{y} \leftarrow \text{Classifier}(\bar{z})$ & 
Classification head &
$(B, d)$ &
$(B, M)$ &
\texttt{torch.Linear}$^{1}$ \\

\hline
\end{tabular}
\end{adjustbox}
\caption{Step-by-step pseudo code of CT-SSG with semantic description, tensor shapes, and python modules. $^{1}$ refers to {\tt PyTorch} module. $^{2}$ refers to {\tt torchvision} module. $^{3}$ refers to {\tt PyTorch Geometric} module.}
\label{appendix:table:pseudo_code}
\end{table*}

Table~\ref{appendix:table:pseudo_code} presents a step-by-step pseudo code for CT-SSG implementation.

%%%%%%%%%%%%%%%%%%%%%%%%%%%
% t-SNE

\section{t-SNE visualization} \label{sec:appendix:tsne}

\begin{figure*}[h]
	\centering
	\includegraphics[width=0.95\textwidth]{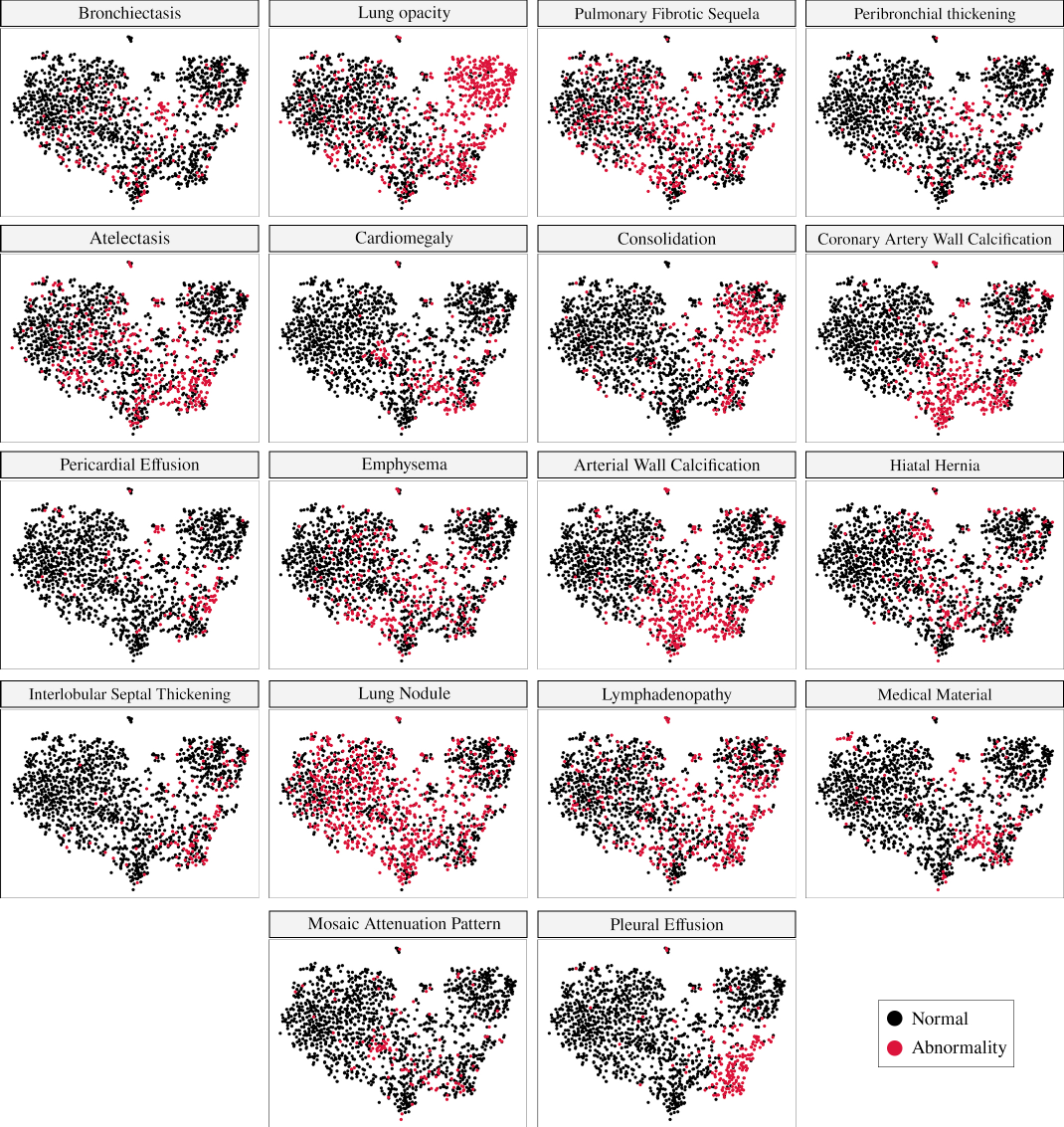}
	\caption{t-SNE visualization, of the pooled features for the {\tt CT-RATE} test dataset. The colors represent classes.}
    \label{fig:tsne}
\end{figure*}

Figure~\ref{fig:tsne} illustrates a t-SNE visualization of the pooled features vector, denoted as $\bar{z}$ and defined in Equation~\ref{equation:pooled_features}. The t-SNE was implemented using \href{https://scikit-learn.org/stable/modules/generated/sklearn.manifold.TSNE.html#sklearn.manifold.TSNE}{scikit-learn t-SNE module}, with a $2$-dimensional latent space and a perplexity of $30$.

\section{Comparison with Generalist Computed Tomography Foundation Models} \label{sec:appendix:foundation}

Table~\ref{appendix:table:comparison_foundation} summarizes the configurations of the evaluated CT-specific foundation models under the linear probing setting. For each model, latent representations are extracted from pretrained and frozen visual encoders. Preprocessing follows the protocols reported in the corresponding original works, including volume orientation, spatial resolution, and Hounsfield Unit (HU) range, to ensure a fair comparison. To further contextualize multi-label abnormality classification performance, the table also reports pretraining objectives, training datasets and their sizes, anatomical coverage, and the latent dimensionality of the backbone representations.

\begin{table*}[h]
\centering
\begin{adjustbox}{width=1.00\textwidth}
\begin{tabular}{c l l l l l l l l l c c}
\toprule

\multicolumn{3}{c}{Method} & \multicolumn{3}{c}{Data} & \multicolumn{2}{c}{Processing} & \multicolumn{2}{c}{Complexity} & \multicolumn{2}{c}{\texttt{RAD-ChestCT} Metrics}\\
\cmidrule(lr){1-3} \cmidrule(lr){4-6} \cmidrule(lr){7-8} \cmidrule(lr){9-10} \cmidrule(lr){11-12}

Setup & Model & Pretrain task(s) & Dataset(s) & Train size & Region(s) & Spacing (mm) & HU range & Param. & Dim. & AUROC & mAP\\

\midrule

%%%%%%%%%%%%%%%%%%%%%%%%%%%
% Foundation models
%%%%%%%%%%%%%%%%%%%%%%%%%%%

% CT-FM
\multirow{4}{*}{\rotatebox[origin=c]{0}{\small \begin{tabular}{@{}c@{}}
\small Foundation\\Models\\Probing\\
\end{tabular}}}& 
\cellcolor[gray]{0.925}\textbf{CT-FM} & 
\cellcolor[gray]{0.925}Contrastive &
\cellcolor[gray]{0.925}69 cohorts$^{(1)}$ &
\cellcolor[gray]{0.925}148k &
\cellcolor[gray]{0.925}Chest, Abdomen$^{(2)}$ &
\cellcolor[gray]{0.925}$(3.0, 1.0, 1.0)$ &
\cellcolor[gray]{0.925}$[-1024, 2048]$ &
\cellcolor[gray]{0.925}$77$M &
\cellcolor[gray]{0.925}$512$ &
\cellcolor[gray]{0.925}$63.26 \text{\textcolor{gray}{\scriptsize $\pm 0.13$}}$ &
\cellcolor[gray]{0.925}$43.78 \text{\textcolor{gray}{\scriptsize $\pm 0.15$}}$ \\

% CT-CLIP
&
\textbf{CT-CLIP} & 
Vision-language &
\texttt{CT-RATE} &
20k &
Chest &
$(1.5, .75, .75)$ &
$[-1000, 1000]$ &
$176$M$^{(3)}$ &
$512$ &
$63.38 \text{\textcolor{gray}{\scriptsize $\pm 0.05$}}$ &
$44.30 \text{\textcolor{gray}{\scriptsize $\pm 0.06$}}$ \\

% MERLIN
&
\cellcolor[gray]{0.925}\textbf{Merlin} & 
\cellcolor[gray]{0.925}Vision-language &
\cellcolor[gray]{0.925}\texttt{MerlinAbdCT} &
\cellcolor[gray]{0.925}15k &
\cellcolor[gray]{0.925}Abdomen$^{(4)}$ &
\cellcolor[gray]{0.925}$(3.0, 1.5, 1.5)$ &
\cellcolor[gray]{0.925}$[-1000, 1000]$ &
\cellcolor[gray]{0.925}$121$M &
\cellcolor[gray]{0.925}$2048$ &
\cellcolor[gray]{0.925}$69.80 \text{\textcolor{gray}{\scriptsize $\pm 0.05$}}$ &
\cellcolor[gray]{0.925}$49.84 \text{\textcolor{gray}{\scriptsize $\pm 0.08$}}$ \\

% COLIPRI
&
\textbf{COLIPRI} & 
Vision-language, vision-only$^{(5)}$ &
\texttt{CT-RATE}, \texttt{NLST} &
94k &
Chest &
$(2.0, 2.0, 2.0)$ &
$[-1000, 1000]$ &
$147$M &
$768$ &
$73.98 \text{\textcolor{gray}{\scriptsize $\pm 0.02$}}$ &
$55.45 \text{\textcolor{gray}{\scriptsize $\pm 0.10$}}$ \\

\toprule

% CT-SSG
\small Supervised & 
\textbf{CT-SSG} & 
None &
- &
- &
- &
$(1.5, .75, .75)$ &
$[-1000, 200]$ &
$12$M &
$512$ &
$74.58 \text{\textcolor{gray}{\scriptsize $\pm 0.36$}}$ &
$58.75 \text{\textcolor{gray}{\scriptsize $\pm 0.28$}}$ \\

\bottomrule
\end{tabular}
\end{adjustbox}
%\vspace{-0.8em}
\caption{
Performance comparison of multi-label abnormality classification between CT-SSG (supervised training) and CT-specific foundation models (linear probing),  on \texttt{RAD-ChestCT}. \textit{HU range} denotes the Hounsfield Unit range, \textit{Param.} the number of visual encoder parameters (in millions, M), and \textit{Dim.} the dimensionality of the latent representations.
\newline
\footnotesize$^{(1)}$ The 69 cohorts are detailed in Table S5 of the CT-FM paper.
\newline
$^{(2)}$ Head and neck, Pelvis and Extremity body parts are also included.
\newline
$^{(3)}$ The CT-CLIP visual encoder comprises 26M parameters, with an additional 150M parameters in the visual projection head, which maps flattened feature representations to a $512$-dimensional embedding space.
\newline
$^{(4)}$ Pelvis body parts are also included.
\newline
$^{(5)}$ Report generation and opposite sentence losses are also included.
}
\label{appendix:table:comparison_foundation}
\end{table*}

\section{Training protocols}

Table~\ref{appendix:table:summary_protocol} summaries backbone initialization and training protocols for all methods. Table~\ref{table:appendix:baseline_implementation} provides links to the baseline implementations and model weights for initialization or direct comparison.

% TABLE METRICS WITH ALL LABELS
\renewcommand{\arraystretch}{1.5}
\begin{table*}[t]
\centering
\begin{adjustbox}{width=1.0\textwidth}
\begin{tabular}{l c c c c c c c c}
\hline
\textbf{Component} & \textbf{ResNet3D} & \textbf{ViT3D} & \textbf{ViViT} & \textbf{Swin3D} & \textbf{CT-Net} & \textbf{CT-MvG} & \textbf{CT-Scroll} & \textbf{CT-SSG} \\
\hline

%%%%%%%%%%%%%
Optimizer & 
Adam &
Adam &
Adam &
Adam &
Adam &
Adam &
Adam &
Adam \\ % 

%%%%%%%%%%%%%%
\rowcolor[gray]{0.9}
Learning rate &
0.0001 &
0.0001 &
0.0001 &
0.0001 &
0.0001 &
0.0001 &
0.0001 &
0.0001 \\ % SSG

%%%%%%%%%%%%%
$(\beta_{1}, \beta_{2})$ & 
$(0.9, 0.99)$ &
$(0.9, 0.99)$ &
$(0.9, 0.99)$ &
$(0.9, 0.99)$ &
$(0.9, 0.99)$ &
$(0.9, 0.99)$ &
$(0.9, 0.99)$ &
$(0.9, 0.99)$ \\ % 

%%%%%%%%%%%%%%
\rowcolor[gray]{0.9}
Batch size &
4 &
4 &
4 &
4 &
4 &
4 &
4 &
4 \\ %

%%%%%%%%%%%%%
Warm-up steps & 
10,000 &
10,000 &
10,000 &
10,000 &
10,000 &
10,000 &
10,000 &
10,000 \\ % 

%%%%%%%%%%%%%
\rowcolor[gray]{0.9}
Maximum steps & 
200,000 &
200,000 &
200,000 &
200,000 &
200,000 &
200,000 &
200,000 &
200,000 \\ % 

%%%%%%%%%%%%%%
Frozen parameter(s) &
None &
None &
None &
None &
None &
None &
None &
None \\ %

%%%%%%%%%%%%%%
\rowcolor[gray]{0.9}
Weights initialization type &
Inflation &
Inflation &
Inflation &
Inflation &
Direct &
Direct &
Direct &
Direct \\ %

%%%%%%%%%%%%%%
Weights initialization backbone &
ResNet-18 &
ViT-S16 &
ViT-S16 &
ViT-S16 &
ResNet-18 &
ViT-S16 &
ResNet-18 &
ResNet-18 \\ %

%%%%%%%%%%%%%%
\rowcolor[gray]{0.9}
Weights initialization source &
ImageNet &
ImageNet &
ImageNet &
ImageNet &
ImageNet &
ImageNet &
ImageNet &
ImageNet \\ %

\hline
\end{tabular}
\end{adjustbox}
\caption{
Summary of training protocols for all methods.
}
\label{appendix:table:summary_protocol}
\end{table*}

\begin{table*}[h]
\centering
\begin{adjustbox}{width=1.00\textwidth}
\begin{tabular}{l l l}
\toprule
Method & Implementation & Model weights\\
\toprule

ResNet3D &
\url{https://github.com/PPPrior/i3d-pytorch} &
\url{https://download.pytorch.org/models/resnet18-f37072fd.pth} \\

\cellcolor[gray]{0.925}ViT3D &
\cellcolor[gray]{0.925}\url{https://github.com/lucidrains/vit-pytorch} &
\cellcolor[gray]{0.925}\url{https://huggingface.co/google/vit-base-patch16-224} \\

ViViT &
\url{https://github.com/lucidrains/vit-pytorch} &
\url{https://huggingface.co/google/vit-base-patch16-224} \\

\cellcolor[gray]{0.925}Swin3D &
\cellcolor[gray]{0.925}\url{https://github.com/microsoft/Swin3D} &
\cellcolor[gray]{0.925}\url{https://huggingface.co/microsoft/swin-small-patch4-window7-224} \\

CT-Net &
\url{https://github.com/rachellea/ct-net-models} &
Not available \\

\cellcolor[gray]{0.925}CT-MvG &
\cellcolor[gray]{0.925}\url{https://github.com/compai-lab/2024-miccai-grail-kiechle} &
\cellcolor[gray]{0.925}Not available \\

CT-Scroll &
\url{https://github.com/theodpzz/ct-scroll} &
\url{https://huggingface.co/theodpzz/ct-scroll} \\

\bottomrule

\cellcolor[gray]{0.925}CT-FM &
\cellcolor[gray]{0.925}\url{https://github.com/project-lighter/CT-FM} &
\cellcolor[gray]{0.925}\url{https://huggingface.co/project-lighter/ct_fm_feature_extractor} \\

CT-CLIP &
\url{https://github.com/ibrahimethemhamamci/CT-CLIP} &
\url{https://huggingface.co/datasets/ibrahimhamamci/CT-RATE} \\

\cellcolor[gray]{0.925}Merlin &
\cellcolor[gray]{0.925}\url{https://github.com/StanfordMIMI/Merlin} &
\cellcolor[gray]{0.925}\url{https://huggingface.co/stanfordmimi/Merlin} \\

COLIPRI &
\url{https://huggingface.co/microsoft/colipri} &
\url{https://huggingface.co/microsoft/colipri} \\

\toprule
\end{tabular}
\end{adjustbox}
%\vspace{-0.8em}
\caption{
Baseline PyTorch implementations and model weights for initialization or direct comparison for backbone initialization, inflation or direct comparison.
}
\label{table:appendix:baseline_implementation}
\end{table*}

% --- Overlap slices ---
\section{Effect of Slice Overlap in 2.5D Graph Construction}

To analyze the effect of overlapping versus non-overlapping axial slice triplets in the proposed framework, Table~\ref{table:appendix:overlap} reports the number of graph nodes, training and inference times, and multi-label abnormality classification performance for both configurations. The two settings achieve comparable classification performance, indicating that introducing overlap provides no measurable performance benefit. In contrast, the overlapping configuration substantially increases the number of graph nodes as well as the training time to convergence and inference time. Based on this performance-efficiency trade-off, the non-overlapping configuration was adopted in the main experiments.

\begin{table}[h]
\centering
\begin{adjustbox}{width=1.00\columnwidth}
\begin{tabular}{c c c c c c}
\toprule
Overlap. slices & Nodes & Train & Inference & F1-Score & AUROC\\
\toprule

% Overlap
$2$ &
$119$ &
24 h &
$105 \text{\textcolor{gray}{\scriptsize $\pm 0.5$}}$ ms&
$57.12 \text{\textcolor{gray}{\scriptsize $\pm 0.21$}}$ &
$\underline{83.72} \text{\textcolor{gray}{\scriptsize $\pm 0.23$}}$  \\

% Non-overlap
None &
$80$ &
18 h &
$70 \text{\textcolor{gray}{\scriptsize $\pm 0.5$}}$ ms&
$\underline{57.18} \text{\textcolor{gray}{\scriptsize $\pm 0.19$}}$ &
$83.64 \text{\textcolor{gray}{\scriptsize $\pm 0.21$}}$ \\

\toprule
\end{tabular}
\end{adjustbox}
%\vspace{-0.8em}
\caption{
Performance comparison between non-overlapping and overlapping triplets of axial slices. \textit{Overlap. slices} refers to the number of overlapping axial slices between two consecutive nodes. \textit{Train} is the training time to reach convergence (in hours, h) and \textit{Inference} the inference time for a sample (in milliseconds, ms) estimated on a NVIDIA RTX A6000 GPU. The two configurations achieve comparable performance, but the overlapping one increases both the training and inference times.
}
\label{table:appendix:overlap}
\end{table}

% --- Scaling 2.5D backbone ---
\section{2D Backbone scaling}
\label{sec:appendix:scaling}

While our work primarily focuses on graph-based feature aggregation and deliberately adopts ResNet-18 as an efficient 2.5D backbone, we evaluate the scalability of CT-SSG using a higher-capacity encoder. To ensure a fair comparison, we replace the ResNet-18 backbone with ResNet-34 while preserving all other architectural and training settings. Table~\ref{table:appendix:encoder_scaling} reports parameter counts and performance obtained with both encoders. The results demonstrate that CT-SSG benefits from increased backbone capacity, highlighting the backbone-agnostic design of CT-SSG and its ability to scale with stronger feature extractors.

\begin{table}[h]
\centering
\begin{adjustbox}{width=1.00\columnwidth}
\begin{tabular}{l c c c c c}

\midrule

\multicolumn{2}{c}{\texttt{Model}} & \multicolumn{4}{c}{Metrics}  \\
\cmidrule(lr){1-2} \cmidrule(lr){3-6}

Backbone & Params & F1-Score & AUROC & mAP & Accuracy\\

\midrule

\rowcolor{gray!10}
\multicolumn{6}{c}{{\texttt{CT-RATE}}} \\

% ResNet18
ResNet18 &
12M &
$57.18 \text{\textcolor{gray}{\scriptsize $\pm 0.19$}}$ &
$83.64 \text{\textcolor{gray}{\scriptsize $\pm 0.21$}}$ &
$58.24 \text{\textcolor{gray}{\scriptsize $\pm 0.50$}}$ &
$81.03 \text{\textcolor{gray}{\scriptsize $\pm 0.38$}}$  \\

% ResNet34
ResNet34 &
22M &
$\underline{57.84} \text{\textcolor{gray}{\scriptsize $\pm 0.21$}}$ &
$\underline{83.72} \text{\textcolor{gray}{\scriptsize $\pm 0.23$}}$ &
$\underline{58.53} \text{\textcolor{gray}{\scriptsize $\pm 0.49$}}$ &
$\underline{81.36} \text{\textcolor{gray}{\scriptsize $\pm 0.32$}}$  \\

%%%%%%%%%%%%%%%
% RAD-ChestCT
\midrule
\rowcolor{gray!10}
\multicolumn{6}{c}{{\texttt{RAD-ChestCT}}} \\

% ResNet18
ResNet18 &
12M &
$52.25 \text{\textcolor{gray}{\scriptsize $\pm 0.88$}}$ &
$74.58 \text{\textcolor{gray}{\scriptsize $\pm 0.36$}}$ &
$58.75 \text{\textcolor{gray}{\scriptsize $\pm 0.28$}}$ &
$69.37 \text{\textcolor{gray}{\scriptsize $\pm 1.74$}}$  \\

% ResNet34
ResNet34 &
22M &
$\underline{52.48} \text{\textcolor{gray}{\scriptsize $\pm 0.22$}}$ &
$\underline{75.60} \text{\textcolor{gray}{\scriptsize $\pm 0.31$}}$ &
$\underline{58.90} \text{\textcolor{gray}{\scriptsize $\pm 0.26$}}$ &
$\underline{70.76} \text{\textcolor{gray}{\scriptsize $\pm 0.28$}}$ \\

\toprule
\end{tabular}
\end{adjustbox}
%\vspace{-0.8em}
\caption{
Comparison of CT-SSG performance using both ResNet-18 and ResNet-34 encoders. The table reports model parameter counts in millions (\textit{Params}, M) and classification performance when node features are extracted using a backbone of increasing capacity, while keeping all other architectural and training settings unchanged. Results show that CT-SSG can benefit from stronger feature encoders, supporting the backbone-agnostic and scalable design of the proposed framework.
}
\label{table:appendix:encoder_scaling}
\end{table}

% --- Effect of the number of slices per nodes ---
\section{Number of axial slices per node}

CT-SSG design models the CT Scan as a graph of nodes that correspond to triplets of axial slices, allowing to capture inter-slice anatomical continuity and subtle volumetric variations while preserving high in-plane resolution by matching the 3 channel RGB structure of 2D backbones for initial feature extraction. To analyze how the number of axial slices per node impact performance, we further investigate CT-SSG performance with 1 unique axial slice per node and 6 unique slices per node. To ensure fair comparison across configurations, we conserve the same CT-SSG's architectural components and extract initial node features by inflating the 2D ResNet backbone weights into 3D, following~\cite{carreira_quo_2018}, a strategy that enables to vary the number of axial slices encoded within a node. Table~\ref{table:appendix:slices_per_node} reports multi-label abnormality classification performance for 1, 3 and 6 adjacent axial slices per node. CT-SSG yields the best results with 3 axial slices per node, whose initial features are extracted by a 2D ResNet. We observe that CT-SSG yields the best performance with 3 axial slices per node, achieving a favourable balance between local anatomical continuity and representations compactness.

\begin{table}[h]
\centering
\begin{adjustbox}{width=1.00\columnwidth}
\begin{tabular}{c c c c c c}
\toprule
Slices per node & Nodes & F1-Score & AUROC & mAP \\
\toprule

% 1 slice per node
1 &
240 &
$56.64 \text{\textcolor{gray}{\scriptsize $\pm 0.26$}}$ &
$83.34 \text{\textcolor{gray}{\scriptsize $\pm 0.29$}}$ &
$57.31 \text{\textcolor{gray}{\scriptsize $\pm 0.25$}}$ \\

% 3 slice per node
3 &
80 &
$\underline{57.18} \text{\textcolor{gray}{\scriptsize $\pm 0.19$}}$ &
$\underline{83.64} \text{\textcolor{gray}{\scriptsize $\pm 0.21$}}$ &
$\underline{58.24} \text{\textcolor{gray}{\scriptsize $\pm 0.50$}}$ \\

% 6 slices per node
6 &
40 &
$55.24 \text{\textcolor{gray}{\scriptsize $\pm 0.30$}}$ &
$82.22 \text{\textcolor{gray}{\scriptsize $\pm 0.07$}}$ &
$55.54 \text{\textcolor{gray}{\scriptsize $\pm 0.10$}}$ \\

\toprule
\end{tabular}
\end{adjustbox}
%\vspace{-0.8em}
\caption{
Comparison of CT-SSG performance for different number of axial slices per node. \textit{Slices per node} is the number of adjacent axial slices per node, and \textit{Nodes} is the number of nodes in our graph construction. CT-SSG yields best performance with 3 slices per node, where initial node features are extracted by a 2D ResNet.}
\label{table:appendix:slices_per_node}
\end{table}

\end{document}